\documentclass[10pt,twocolumn,letterpaper]{article}

\usepackage[pagenumbers]{cvpr} 

\definecolor{cvprblue}{rgb}{0.21,0.49,0.74}
\usepackage[pagebackref,breaklinks,colorlinks,allcolors=cvprblue]{hyperref}

\usepackage[utf8]{inputenc}
\usepackage[T1]{fontenc}
\usepackage{hyperref}
\usepackage{url}     
\usepackage{booktabs}
\usepackage{amsfonts}
\usepackage{nicefrac}
\usepackage{microtype}

\usepackage{graphicx}
\usepackage{booktabs}
\usepackage{multicol,multirow}
\usepackage{amssymb}
\usepackage{comment}
\usepackage[dvipsnames]{xcolor}
\usepackage{pifont}
\usepackage[ruled,vlined]{algorithm2e}
\usepackage{algpseudocode}
\usepackage{amsmath}
\usepackage{enumitem}
\usepackage{placeins}
\usepackage{caption}
\usepackage{wrapfig}

\newcommand{\cmark}{{\color{deltaup}\ding{51}}}
\newcommand{\xmark}{{\color{deltadn}\ding{55}}}
\definecolor{deltaup}{RGB}{0,120,60}
\definecolor{deltadn}{RGB}{180,30,30}

\def\confName{CVPR}
\def\confYear{2027}

\title{
Who Did What to Whom and Where Is It Grounded? \\
A Benchmark for Instance-Grounded Video Anomaly Understanding
}

\author{
Inpyo Song$^{1}$ \quad
Eunji Chon$^{1}$ \quad
Joonhyung Kwon$^{1}$ \quad
Minjun Ju$^{2}$ \quad
Jangwon Lee$^{1}$\\
$^{1}$Sungkyunkwan University \quad
$^{2}$Korean Air
}

\makeatletter
\g@addto@macro\@maketitle{%
  \begin{center}
    \begin{minipage}{\textwidth}
      \centering
      \includegraphics[width=0.95\linewidth]{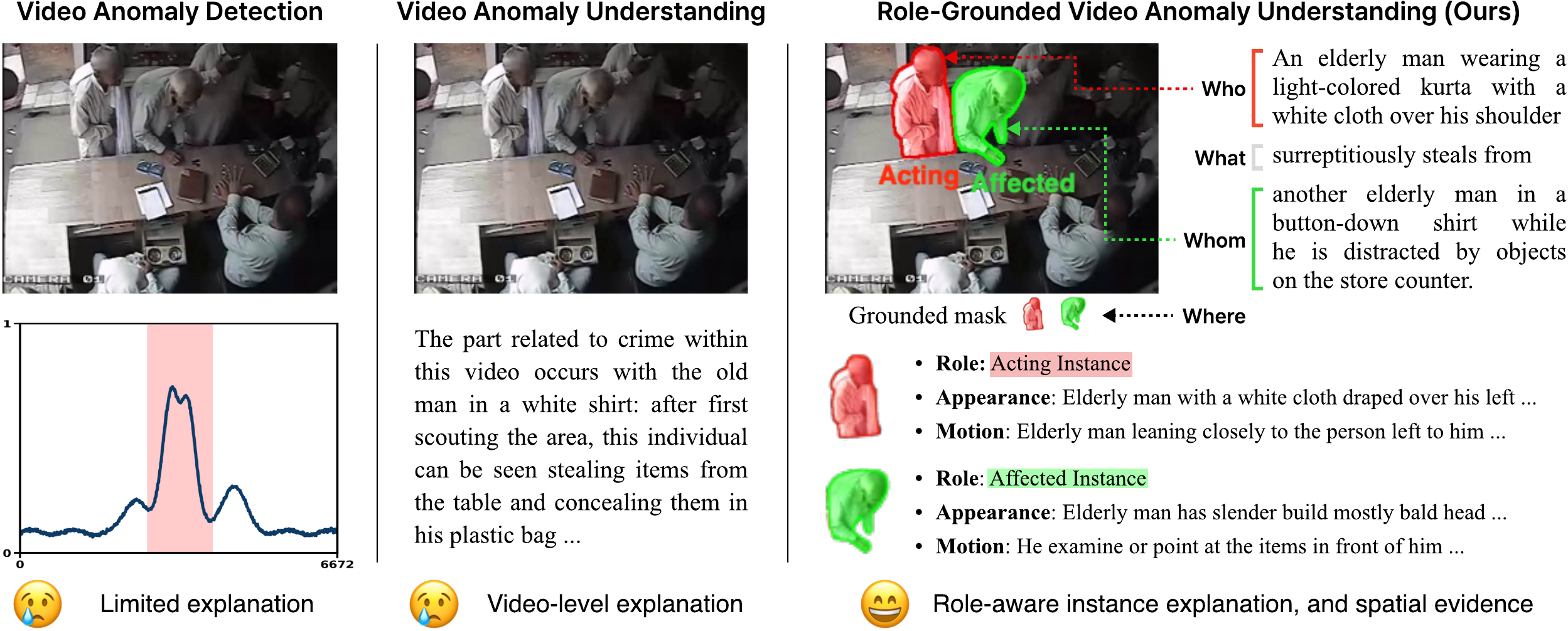}
      \captionsetup{hypcap=false}
      \captionof{figure}{
      From anomaly detection to instance-grounded explanation.
      \textbf{Left}: anomaly scores indicate anomalous periods without describing the event.
      \textbf{Middle}: a video-level description narrates the event without linking its references to specific instances.
      \textbf{Right}: our benchmark links an anomaly caption to acting (red) and affected (green) role-group masks and provides appearance and motion captions for each instance.
      Matching colors indicate the correspondence between captions and masks.
      }
      \label{fig:fig1_front}
    \end{minipage}
  \end{center}
}
\makeatother

\begin{document}
\maketitle

\begin{abstract}
Despite recent advances in video anomaly detection (VAD),
explanations often remain unverifiable because they do not link their claims to visual evidence,
particularly in multi-instance interactions.
To bridge this gap, we introduce AnomGround, a benchmark for instance-grounded video anomaly understanding.
Our annotations explicitly distinguish
who acted in the anomalous event,
what action transpired,
whom it affected,
and where the evidence lies.
Across six VAD datasets, AnomGround provides human-reviewed captions and mask tracks.
Anomaly captions link to role-group tracks, while appearance and motion captions link to individual tracks.
We jointly evaluate caption quality, role-group localization, and grounding of each participant in its declared role.
Joint evaluation exposes errors hidden by caption similarity alone.
Models can describe anomalies accurately yet fail to ground affected participants or link captions to the correct instances.
The annotations will be publicly released.
\end{abstract}

\section{Introduction}
\label{sec:intro}
Explainable artificial intelligence has become a cornerstone of modern machine learning applications,
enabling users to understand model decisions and evaluate whether those decisions can be trusted~\citep{rudin2019stop, mersha2024explainable}.
This capability is particularly important in safety-critical applications such as medical diagnosis and autonomous driving, where model outputs inform decisions that affect human safety~\citep{arun2021assessing-medical, sun2024pathmmu, song2025real}.
In video anomaly detection (VAD), explanations can help users distinguish false positives from genuine threats and provide actionable insights.
Recent VAD research thus increasingly uses large language models (LLMs) and vision-language models (VLMs) to complement anomaly scores and heatmaps with textual descriptions of events~\citep{song2024video, lv2024video, yang2024follow, tang2024hawk, ye2025vera}.

\begin{table*}[t]
  \centering
  \caption{Comparison of released annotations across VAD datasets, text-augmented benchmarks, and this work.
  \textbf{Spatial ann.} denotes released bounding-box (BBox) or mask annotations.
  Instance track denotes whether per-instance temporal tracks are provided.
  \textbf{Text ann.} denotes released text annotations, including question answering (QA).
  \textbf{Role ann.} denotes whether acting and affected instance roles are explicitly labeled.
  \textbf{Text-to-track} denotes explicit links from text to spatial tracks.}
  \label{tab:dataset_comparison}
  \setlength{\tabcolsep}{3.5pt}
  \scriptsize
  \resizebox{0.9\linewidth}{!}{%
  \begin{tabular}{lccccc}
    \toprule
    \textbf{Benchmark} & \textbf{Spatial ann.} & \textbf{Instance track} & \textbf{Text ann.} & \textbf{Role ann.} & \textbf{Text-to-track} \\
    \midrule
    \multicolumn{6}{l}{\textbf{VAD datasets}} \\
    Ped1~\citep{li2013anomaly} & Mask (subset) & \xmark & \xmark & \xmark & \xmark \\
    Ped2~\citep{li2013anomaly} & Mask (subset) & \xmark & \xmark & \xmark & \xmark \\
    Avenue~\citep{lu2013abnormal} & BBox & \xmark & \xmark & \xmark & \xmark \\
    ShanghaiTech~\citep{liu2018future} & Mask & \xmark & \xmark & \xmark & \xmark \\
    UBnormal~\citep{acsintoae2022ubnormal} & Mask & \xmark & \xmark & \xmark & \xmark \\
    UCF-Crime~\citep{sultani2018real} & \xmark & \xmark & \xmark & \xmark & \xmark \\
    VIEW360~\citep{song2025anomaly} & Direction & \xmark & \xmark & \xmark & \xmark \\
    MSAD~\citep{msad} & \xmark & \xmark & \xmark & \xmark & \xmark \\
    \midrule
    \multicolumn{6}{l}{\textbf{Text-augmented benchmarks}} \\
    UCF-CAP~\citep{chatzikonstantinou2022ucfcap} & \xmark & \xmark & Caption & \xmark & \xmark \\
    UCA~\citep{yuan2024uca} & \xmark & \xmark & Caption & \xmark & \xmark \\
    UCFCrime-AR~\citep{wu2024var} & \xmark & \xmark & Caption & \xmark & \xmark \\
    HAWK~\citep{tang2024hawk} & \xmark & \xmark & Caption, QA & \xmark & \xmark \\
    CUVA~\citep{du2024uncovering} & \xmark & \xmark & QA & \xmark & \xmark \\
    UCCD~\citep{zhou2024uccd} & BBox (Person) & \xmark & Caption & \xmark & \xmark \\
    VAD-Instruct50k~\citep{zhang2024holmes} & \xmark & \xmark & Caption, Instruction & \xmark & \xmark \\
    HIVAU-70k~\citep{zhang2025holmesvau} & \xmark & \xmark & Caption, Instruction & \xmark & \xmark \\
    VAGU~\citep{gao2025vagu} & \xmark & \xmark & Explanation, QA & \xmark & \xmark \\
    ECVA~\citep{du2024ecva} & \xmark & \xmark & QA & \xmark & \xmark \\
    \midrule
    \textbf{AnomGround} & Mask & \cmark & Caption & Acting \& Affected & \cmark \\
    \bottomrule
  \end{tabular}
  }
\end{table*}

However, textual descriptions can leave unclear which visual regions support the explanation, 
despite they improve interpretability over score-only outputs.
Figure~\ref{fig:fig1_front} (left) shows how anomaly scores indicate when an event may occur without describing its participants.
Figure~\ref{fig:fig1_front} (middle) offers a richer description, 
yet ``the old man in a white shirt'' does not clearly distinguish among men wearing similar clothing and leaves the person affected by the theft unspecified.
Such descriptions can remain ambiguous and incomplete~\citep{wang2024videohallucer}, 
and the absence of explicit spatial links makes their visual support difficult to verify~\citep{li2022invariant,Li_2023_ICCV}.
These gaps leave open four questions:
\textbf{1.~who} performed the anomalous action,
\textbf{2.~what} each instance was doing,
\textbf{3.~whom} the event affected, and
\textbf{4.~where} the explanation is grounded in the video.

To address these limitations, we introduce \textbf{AnomGround}, a benchmark for instance-grounded video anomaly understanding across six established VAD datasets.
Anomaly captions describe the event and link acting and affected phrases to role-group mask tracks, each combining the masks of participants that share a role.
Appearance and motion captions describe each participant's visible attributes and behavior, paired with its individual mask track.
Figure~\ref{fig:fig1_front} (right) illustrates how these annotations connect the event to its participants and their visual evidence.
Every mask frame, role label, and caption undergoes human review and correction across pedestrian, campus, traffic, and crime scenarios.
Table~\ref{tab:dataset_comparison} shows that the compared benchmarks lack this combination of role labels, individual mask tracks, and explicit links between captions and tracks.

These annotations support joint evaluation of descriptions and their spatial evidence.
Following prior region-caption evaluations~\citep{johnson2016densecap, chen2021scan2cap}, we also introduce a grounding-conditioned caption score (GCC), which credits caption similarity only when the associated tracks meet the localization criterion.
Anomaly captions are evaluated against their role groups, while appearance and motion captions are evaluated against individual tracks.
For anomaly captions, role-aware GCC (rGCC) evaluates both the declared role-group tracks and the localization of every annotated member.
A small participant with inadequate local support can therefore affect the score even when its role group has substantial overlap.

Our evaluation of grounded VLMs and VLMs coupled to Segment Anything Model 2 (SAM2)~\citep{ravi2024sam} reveals errors hidden by caption quality alone.
Among the VLM+SAM2 pipelines, joint evaluation reverses a model ordering obtained from anomaly caption similarity.
Group overlap can also conceal inadequate localization of individual participants.
Role-specific analysis also shows lower appearance and motion joint scores for affected participants across both model groups.
AnomGround makes these differences measurable by linking event descriptions to the roles and individual tracks of their participants.

\noindent Our contributions are:
\begin{itemize}[
    topsep=0pt,
    partopsep=0pt,
    itemsep=0.5em,
    parsep=0pt,
    leftmargin=*
]
    \item \textbf{An instance-grounded anomaly benchmark.}
    AnomGround connects event descriptions, participant roles, and individual appearance and motion captions through mask tracks across six VAD datasets.
    Every mask frame and caption receives human review and correction.
    \item \textbf{Joint caption and participant evaluation.}
    GCC evaluates descriptions with their spatial support, while rGCC checks every annotated participant within its declared role, including members of a collectively described group.
    \item \textbf{Insights for explainable VAD.}
    Joint evaluation reveals that strong captioning performance does not ensure correct grounding of the described instances.
\end{itemize}

\section{Related Work}
\label{sec:related_work}

\paragraph{Explainable and text-augmented VAD.}
Video anomaly detection is typically studied in semi-supervised or weakly supervised settings that learn normality from normal videos or video-level labels~\citep{hasan2016learning, liu2018future, sultani2018real, tian2021weakly}.
Explainable VAD connects detection decisions to semantic concepts, interpretable rationales, and descriptions generated by LLMs and VLMs.
Recent work uses rule-based reasoning, caption-guided detection, causal explanations, and open-world formulations~\citep{hinami2017joint, doshi2023towards, wu2022explainable, singh2024tracklet, yang2024follow, lv2024video, zhang2024holmes, du2024uncovering, tang2024hawk, ye2025vera, xu2025plovad}.
In parallel, text-augmented benchmarks add video-level captions, question-answer pairs, causal explanations, and instruction-tuning supervision~\citep{chatzikonstantinou2022ucfcap, yuan2024uca, wu2024var, du2024uncovering, tang2024hawk, gao2025vagu, du2024ecva, zhang2024holmes, song2024video, zhang2025holmesvau}.
At the instance level, UCCD annotates tracked first-frame boxes with per-person captions~\citep{zhou2024uccd}.
However, these efforts remain largely video-level or box-level, without human-validated per-frame mask tracks, 
explicit acting/affected role labels, or direct links between textual claims and spatial evidence.

\paragraph{Visual-text grounding.}
Grounded video understanding has shown that fluent language or correct answers do not guarantee alignment with the right visual evidence~\citep{li2022invariant, Li_2023_ICCV, xiao2024can}.
Related tasks localize language-described instances with per-frame masks, and recent multimodal models increasingly combine dense grounding with language generation~\citep{botach2022end, wu2022language, wu2023onlinerefer, ding2025mevis, lai2024lisa, bai2024videolisa, rasheed2024glamm, munasinghe2025videoglamm, yuan2025sa2va}.
Object trajectory captioning further connects descriptions to tracked masks~\citep{fiastre2025maskcaptioner}.
AnomGround evaluates whether anomaly explanations are grounded in the correct spatiotemporal evidence.
We jointly annotate per-frame mask tracks, acting and affected role labels, and text-to-track links, penalizing explanations assigned to the wrong instance (Table~\ref{tab:dataset_comparison}).

\section{The AnomGround Benchmark}
\label{sec:benchmark}

\begin{figure*}[t]
  \centering
  \includegraphics[width=\textwidth]{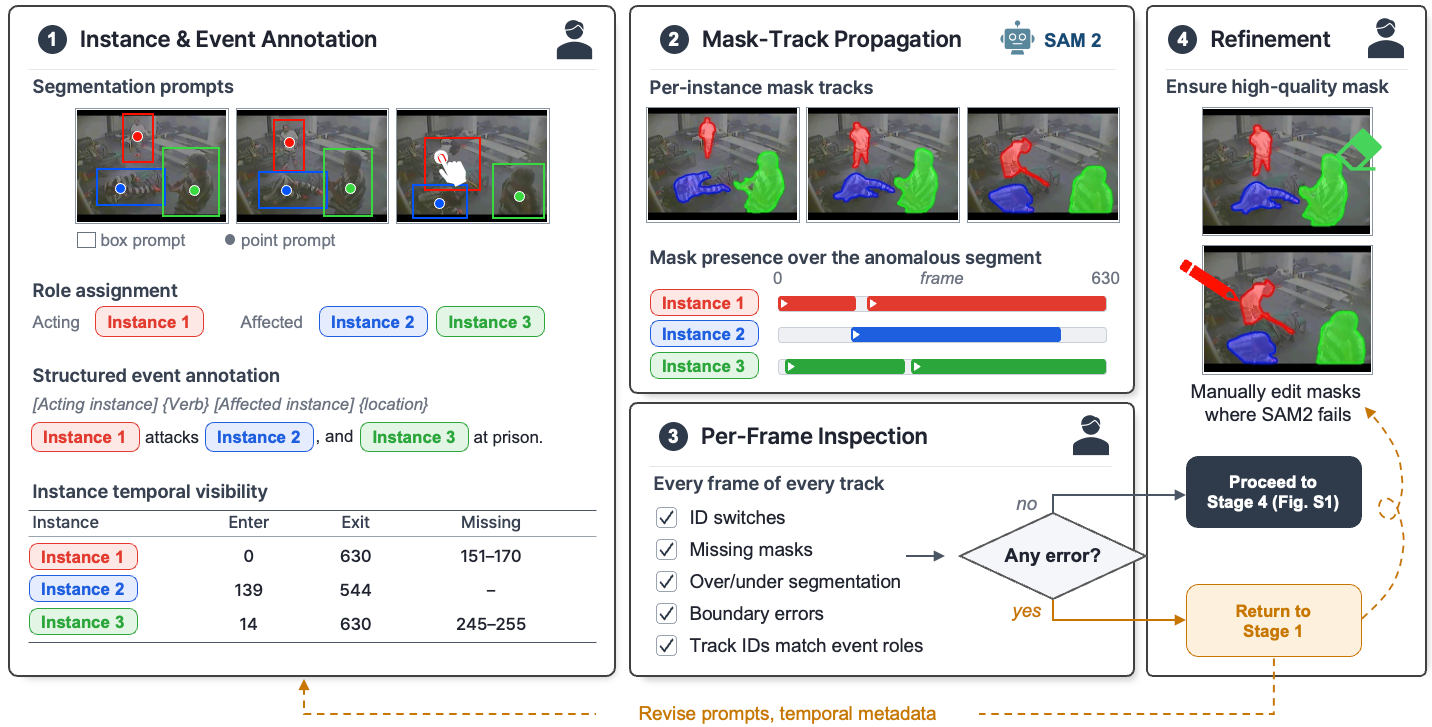}
  \caption{
  \textbf{Mask annotation, inspection, and correction.}
  Annotators assign event roles and spatial prompts to individual instances.
  SAM~2 propagates masks over the annotated temporal spans.
  Every frame of every track is inspected for instance identity, mask quality, and consistency with the event roles.
  Errors are corrected by refining prompts and temporal metadata or by direct pixel-level editing.
  Colors distinguish individual instance tracks.
  The reviewed masks support caption generation in Suppl.~Fig.~\ref{fig:supp_caption_pipeline}.
  }
  \label{fig:mask_annotation_pipeline}
\end{figure*}

\subsection{Task and Annotations}
\label{sec:task_def}

AnomGround evaluates descriptions of anomalous video clips together with the visual instances supporting those descriptions.
Each clip spans an anomaly interval annotated in a source dataset.
Given a clip, a model generates an anomaly caption, appearance and motion captions for its participants, and the corresponding mask tracks.
Ground-truth captions, role labels, and masks are withheld from the model.

At the event level, an \emph{anomaly caption} describes what happened through the actions and interactions of the visible participants.
An \emph{acting instance} performs the anomalous action or forms the visible source of an anomalous process.
An \emph{affected instance} is acted upon or directly affected by that action or process.
These roles apply to the annotated visible instances, so a role is absent when it has no annotated instance.

An anomaly caption can describe several participants through a single role phrase.
For example, in ``three people attack a man,'' the acting phrase refers to three instances and the affected phrase to one.
We match this unit of description with a \emph{role-group mask track}, formed by taking the union of the corresponding instance masks at each frame.
The anomaly caption is paired with the tracks for its acting and affected groups.
The model is prompted to identify acting and affected phrases in its generated anomaly caption and associate each phrase with a mask track.
Tracks assigned to the same role are combined by taking their union at each frame.

At the instance level, an \emph{appearance caption} describes visible attributes such as clothing, shape, and color.
A \emph{motion caption} describes the instance's actions and state changes over time, including its interactions with other participants.
Each instance retains its individual mask track, paired with both captions.
These individual tracks also form the role-group tracks, allowing the same participants to be studied collectively in an event or separately through their attributes and actions.
Figure~\ref{fig:fig1_front} (right) illustrates this connection between the event description, individual captions, and mask tracks.

\subsection{Dataset Coverage}
\label{sec:dataset_overview}

We started from all anomalous videos in six established VAD datasets: Ped1 and Ped2~\citep{li2013anomaly}, Avenue~\citep{lu2013abnormal}, ShanghaiTech~\citep{liu2018future}, UCF-Crime~\citep{sultani2018real}, and MSAD~\citep{msad}.
Their annotated event intervals provide the clips for AnomGround.
The benchmark contains 763 instance tracks, comprising 228,190 frame-level masks, and 1,938 captions across pedestrian, campus, traffic, and crime scenarios.
Mask counts by role are provided in Suppl.~Sec.~\ref{sec:supp_dataset_statistics}.

Table~\ref{tab:dataset_stats} summarizes annotation coverage and participant composition.
Clips with multiple annotated instances occur in every source dataset.
Ped1 and Ped2 contain only acting instances, while UCF-Crime and MSAD have larger proportions of clips with affected instances.
These differences support analyses of grounding performance across participant counts and role compositions, alongside the comparison across scene types.

\begin{table}[t]
\centering
\caption{
Annotation coverage and participant composition.
Masks are counted per instance and frame.
Multi. and Aff. give the percentages of clips with multiple instances and at least one affected instance, respectively.
}
\label{tab:dataset_stats}
\setlength{\tabcolsep}{3pt}
\small
\begin{tabular}{lrrrr}
\toprule
\textbf{Dataset} & \textbf{Masks} & \textbf{Captions} & \textbf{Multi. (\%)} & \textbf{Aff. (\%)} \\
\midrule
Ped1 & 6,929 & 158 & 27.5 & 0.0 \\
Ped2 & 2,718 & 50 & 58.3 & 0.0 \\
Avenue & 7,495 & 125 & 45.2 & 29.0 \\
ShanghaiTech & 30,498 & 442 & 44.2 & 13.5 \\
UCF-Crime & 127,194 & 693 & 74.0 & 58.8 \\
MSAD & 53,356 & 470 & 61.7 & 53.2 \\
\midrule
\textbf{Total} & \textbf{228,190} & \textbf{1,938} & \textbf{56.6} & \textbf{36.4} \\
\bottomrule
\end{tabular}
\end{table}

\subsection{Annotation and Human Review}
\label{sec:annotation}

Figure~\ref{fig:mask_annotation_pipeline} summarizes the mask annotation and review process.
Seven trained annotators identified the event participants, assigned their roles, and recorded event descriptions.
SAM~2~\citep{ravi2024sam} propagated masks from their spatial prompts.
Annotators inspected every frame of each mask track and corrected errors through prompt refinement or direct pixel-level editing.
This process provides dense, manually reviewed mask tracks for surveillance scenes with complex interactions, low resolution, and limited visibility.

\begin{figure}[t]
  \centering
  \includegraphics[width=\columnwidth]{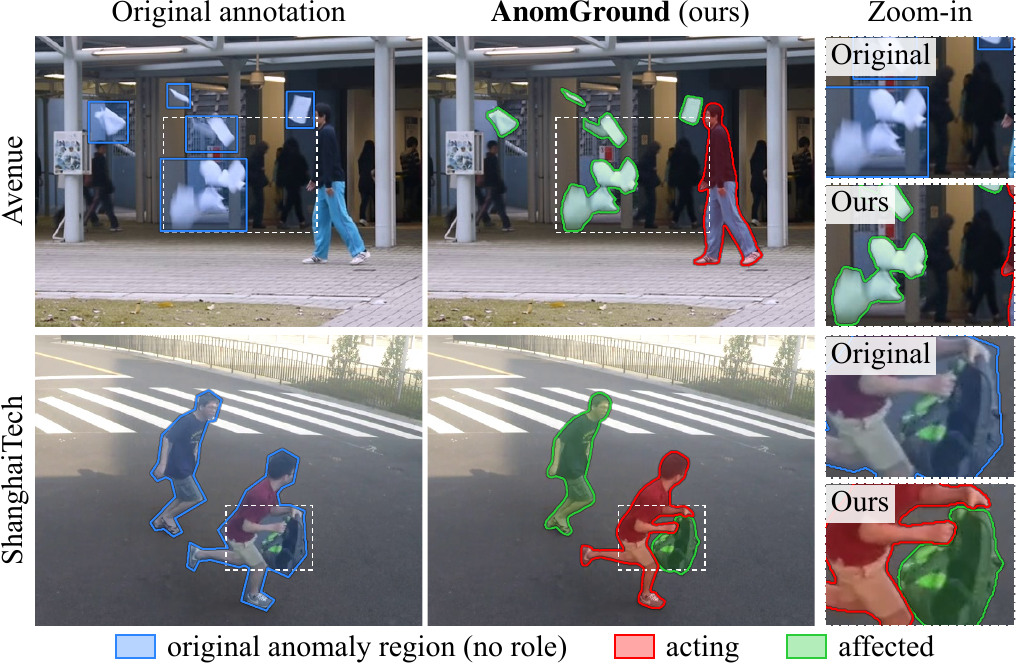}
  \caption{Original anomaly-region annotations and AnomGround instance masks. Our masks follow object boundaries and distinguish acting (red) and affected (green) instances. Insets highlight the boundaries of the scattered papers and the separation of the backpack from the person carrying it.}
  \label{fig:mask_quality}
\end{figure}

AnomGround adds object boundaries and participant roles to the original anomaly-region annotations (Figure~\ref{fig:mask_quality}).
In the paper throwing scene, the masks follow the scattered sheets and distinguish them from the person who throws them.
In the second scene, the masks separate the backpack from the person carrying it, including where the bag and the person's arms overlap.
This spatial detail allows captions about an acting person and an affected object to be evaluated against distinct visual references.

VLMs generated caption drafts from the inspected masks and event annotations.
Annotators reviewed every appearance, motion, and anomaly caption alongside videos with the masks overlaid.
They checked each clip and each instance for caption accuracy and correspondence with the masks, and directly revised the text.
This review covered both the event description and the detailed descriptions of its participants.
Suppl.~Sec.~\ref{sec:supp_benchmark_construction} provides the full procedure and generation prompts.

\section{Evaluation Protocol}
\label{sec:eval}

A caption can describe an event correctly while its mask follows a different participant.
AnomGround evaluates caption content together with its spatial support at the corresponding annotation level: role groups for anomaly captions and individual tracks for appearance and motion.

\subsection{Caption and Mask Evaluation}
\label{sec:matching_scores}

Caption similarity ($\mathrm{Cap}$) measures semantic agreement with the ground-truth description on a scale from zero to one using an LLM judge.
Anomaly Cap compares complete anomaly captions.
For appearance and motion, a one-to-one assignment maximizes total caption similarity independently of mask overlap.

Track intersection-over-union (IoU) averages frame IoU over every annotated frame of the clip.
Following the Densely Annotated Video Segmentation (DAVIS) benchmark convention~\citep{Pont-Tuset_arXiv_2017, Caelles_arXiv_2019}, jointly empty masks receive frame IoU 1, crediting a correct prediction of the instance's absence.
We match predicted and ground-truth tracks by one-to-one Hungarian matching that maximizes total track IoU.
For instance captions, mean IoU (mIoU) averages the matched track IoUs, with zero for unmatched ground-truth instances.
For anomaly GCC, the clip's spatial score is the minimum IoU across its annotated role groups (Sec.~\ref{sec:task_def}), with zero for an unmatched group.
Anomaly mIoU is averaged separately over all annotated acting role-group tracks and all annotated affected role-group tracks.
We also report false-positive instances (FPI), the mean number of unmatched predicted tracks per clip.
Appearance and motion use an IoU threshold of 0.1 to accept a match, while anomaly FPI counts tracks left unmatched by the role-group assignment (Suppl.~Sec.~\ref{sec:supp_reporting}).

\subsection{Grounding-Conditioned Caption Scores}
\label{sec:gcc}

Following prior region-caption joint evaluations~\citep{johnson2016densecap, chen2021scan2cap}, GCC credits a description only when its associated tracks meet an overlap threshold.
For sample $i$, let $\mathcal{M}_i$ contain its required ground-truth tracks and $S_m^{(i)}$ be track $m$'s IoU after matching.
Let $P_i$ equal one when every required track has positive pixel intersection with its matched prediction, and zero otherwise.
For $N$ samples,
\begin{equation}
  \mathrm{GCC}@\tau
  = \frac{1}{N}\sum_{i=1}^{N}
    \mathbf{1}\!\left[\min_{m\in\mathcal{M}_i} S_m^{(i)} \geq \tau\right]
    P_i\,\mathrm{Cap}_i,
  \label{eq:gcc}
\end{equation}
where $\mathrm{Cap}_i$ is the similarity of the evaluated caption to its ground-truth description.
For anomaly captions, each sample is a clip, and every annotated role-group track is required.
This checks each group as a whole, without testing its members separately.
For appearance and motion, each sample is an instance, and $\mathrm{Cap}_i$ scores the caption attached to its IoU-matched prediction, preserving the predicted caption-mask pairing.

Role-aware GCC (rGCC) evaluates role-group outputs at both group and member levels using the declared roles.
Checking each participant can reveal omissions hidden by role-group overlap.
Following local segmentation evaluation~\citep{jaus2025every}, we assign pixels outside a ground-truth role group to the nearest visible instance mask within that role.
Each member's local IoU compares its ground-truth mask with the predicted pixels on that instance and the assigned false-positive pixels.
Other group members are excluded from this local comparison, so errors near a large participant are not charged to every smaller participant.

Member IoUs are averaged over frames in which that ground-truth instance is visible.
Role-group IoUs are averaged over frames with a nonempty ground-truth or predicted group mask.
rGCC retains the anomaly caption score only when every required group and member meets the same IoU threshold.
A missing role declaration or required mask gives zero rGCC.
The spatial assignment and score are defined in Suppl.~Sec.~\ref{sec:supp_reporting}.

We report GCC@0.1 and rGCC@0.1, adopting the overlap threshold used for region matching in Street Scene's region-based and track-based detection criteria~\citep{ramachandra2020street}.
GCC@avg and rGCC@avg average their respective scores over nine thresholds from 0.1 to 0.9 in steps of 0.1.
Cap, mIoU, GCC, and rGCC use pooled-micro averages across the six datasets, over clips for anomaly and ground-truth instances for appearance and motion.
Instance results are also reported separately for acting and affected participants.
Full reporting rules and sensitivity to thresholds and temporal aggregation are provided in Suppl.~Secs.~\ref{sec:supp_reporting}, \ref{sec:supp_robustness}, and~\ref{sec:supp_temporal_aggregation}.

\section{Experimental Results}
\label{sec:experiments}

\begin{table*}[t]
\centering
\caption{Anomaly captioning and grounding on AnomGround.}
\label{tab:main_results}
\small
\setlength{\tabcolsep}{3pt}
\renewcommand{\arraystretch}{1.1}
\begin{tabular*}{\textwidth}{@{\extracolsep{\fill}}lrrrrrrrr@{}}
\toprule
\multirow{2}{*}{\textbf{Model}} & \multirow{2}{*}{\textbf{Cap}$\uparrow$} & \multicolumn{2}{c}{\textbf{mIoU}$\uparrow$} & \multicolumn{2}{c}{\textbf{GCC}$\uparrow$} & \multicolumn{2}{c}{\textbf{rGCC}$\uparrow$} & \multirow{2}{*}{\textbf{FPI}$\downarrow$} \\
\cmidrule(lr){3-4} \cmidrule(lr){5-6} \cmidrule(lr){7-8}
 & & Acting & Affected & @0.1 & @avg & @0.1 & @avg & \\
\midrule
\multicolumn{9}{l}{\textit{Joint caption and mask generation}} \\
Sa2VA-8B & 4.16 & 36.35 & 12.34 & 0.19 & 0.10 & 0.00 & 0.00 & 0.107 \\
VideoGLaMM-3.8B & 2.35 & 26.03 & 16.77 & 0.96 & 0.68 & 0.00 & 0.00 & 0.391 \\
UniPixel-7B & 20.23 & 10.12 & 3.09 & 4.06 & 2.79 & 0.00 & 0.00 & 0.100 \\
\midrule
\multicolumn{9}{l}{\textit{Captioning with separate localization}} \\
Veason-R1-3B & 25.20 & 43.09 & 33.59 & 13.96 & 9.50 & 8.61 & 6.36 & 0.898 \\
InternVL3-8B + SAM2 & 40.87 & 27.35 & 25.78 & 19.54 & 8.12 & 12.15 & 5.56 & 0.575 \\
Qwen3-VL-8B + SAM2 & 43.98 & 50.88 & 38.40 & 33.67 & 20.85 & 24.24 & 16.02 & 0.464 \\
InternVL3.5-8B + SAM2 & 37.26 & 16.34 & 14.93 & 14.31 & 3.88 & 6.76 & 2.01 & 0.502 \\
Qwen3.5-9B + SAM2 & 45.54 & 47.49 & 35.86 & 33.21 & 18.55 & 24.68 & 13.70 & 0.510 \\
\bottomrule
\end{tabular*}
\end{table*}

\begin{table*}[t]
\centering
\caption{Instance captioning and grounding on AnomGround. FPI counts unmatched tracks per clip.}
\label{tab:instance_results}
\small
\setlength{\tabcolsep}{4pt}
\renewcommand{\arraystretch}{1.1}
\begin{tabular*}{\textwidth}{@{\extracolsep{\fill}}lcccccccccc@{}}
\toprule
Model & \multicolumn{5}{c}{\textbf{Appearance}} & \multicolumn{5}{c}{\textbf{Motion}} \\
\cmidrule(lr){2-6} \cmidrule(lr){7-11}
 & Cap$\uparrow$ & mIoU$\uparrow$ & GCC@0.1$\uparrow$ & GCC@avg$\uparrow$ & FPI$\downarrow$ & Cap$\uparrow$ & mIoU$\uparrow$ & GCC@0.1$\uparrow$ & GCC@avg$\uparrow$ & FPI$\downarrow$ \\
\midrule
\multicolumn{11}{l}{\textit{Joint caption and mask generation}} \\
Sa2VA-8B & 4.30 & 27.62 & 3.52 & 2.43 & 0.449 & 6.78 & 31.34 & 5.74 & 3.69 & 0.806 \\
VideoGLaMM-3.8B & 6.87 & 22.40 & 2.60 & 1.87 & 1.223 & 7.49 & 23.41 & 3.49 & 2.47 & 1.148 \\
UniPixel-7B & 1.76 & 2.87 & 1.06 & 0.79 & 0.097 & 13.04 & 15.01 & 5.39 & 3.64 & 1.036 \\
\midrule
\multicolumn{11}{l}{\textit{Captioning with separate localization}} \\
Veason-R1-3B & 15.60 & 26.12 & 11.49 & 9.41 & 0.461 & 20.07 & 24.93 & 12.32 & 8.82 & 0.638 \\
InternVL3-8B + SAM2 & 24.90 & 22.74 & 10.98 & 5.30 & 1.032 & 28.80 & 24.61 & 13.42 & 6.20 & 1.199 \\
Qwen3-VL-8B + SAM2 & 24.91 & 38.40 & 19.31 & 14.60 & 0.425 & 29.85 & 38.33 & 23.98 & 17.20 & 0.444 \\
InternVL3.5-8B + SAM2 & 24.85 & 20.10 & 10.80 & 3.41 & 1.400 & 29.47 & 18.66 & 11.97 & 3.87 & 1.573 \\
Qwen3.5-9B + SAM2 & 37.32 & 44.69 & 27.62 & 19.46 & 1.403 & 33.98 & 43.04 & 26.51 & 17.47 & 1.515 \\
\bottomrule
\end{tabular*}
\end{table*}

\subsection{Setup}
\label{sec:exp_setup}

\emph{Joint caption and mask generation} produces natural-language descriptions and segmentation tokens in the same response.
The model's mask decoder converts these tokens into tracks paired with descriptions in that response.
We evaluate Sa2VA-8B~\citep{yuan2025sa2va}, VideoGLaMM-3.8B~\citep{munasinghe2025videoglamm}, and UniPixel-7B~\citep{liu2026unipixel}.

\emph{Captioning with separate localization} generates captions first, then uses their text to locate participants in subsequent model calls.
Veason-R1-3B~\citep{gong2025reinforcing} takes participant phrases from its caption as queries and predicts a keyframe and boxes through its trained video grounding model.
The VLM+SAM2 pipelines instead predict boxes on the sampled frames using their generated descriptions as context.
Both approaches use SAM~2 to propagate the boxes into mask tracks.
The four VLM+SAM2 backbones are InternVL3-8B~\citep{zhu2025internvl3exploringadvancedtraining}, InternVL3.5-8B~\citep{wang2025internvl35}, Qwen3-VL-8B~\citep{bai2025qwen3vl}, and Qwen3.5-9B~\citep{qwen2026qwen35}.

All models receive the same uniformly sampled frames, up to eight per clip, with masks evaluated over the full clip.
Each model is prompted separately for the three tasks.
For anomaly, a common prompt requests an event sentence with acting and affected participant phrases marked.
We average caption and joint scores over three independent \texttt{gpt-6-luna} judge rounds and report means in the main tables.
All scores except counts use a 100-point scale.
Model settings, prompts, and population standard deviations are provided in Suppl.~Sec.~\ref{sec:supp_eval_baseline}.

\begin{table*}[t]
\centering
\caption{Instance captioning by participant role.}
\label{tab:role_results}
\small
\setlength{\tabcolsep}{1.9pt}
\renewcommand{\arraystretch}{1.1}
\begin{tabular}{lcccccccccccc}
\toprule
Model & \multicolumn{6}{c}{\textbf{Appearance}} & \multicolumn{6}{c}{\textbf{Motion}} \\
\cmidrule(lr){2-7} \cmidrule(lr){8-13}
 & \multicolumn{3}{c}{Acting} & \multicolumn{3}{c}{Affected} & \multicolumn{3}{c}{Acting} & \multicolumn{3}{c}{Affected} \\
\cmidrule(lr){2-4} \cmidrule(lr){5-7} \cmidrule(lr){8-10} \cmidrule(lr){11-13}
 & Cap$\uparrow$ & mIoU$\uparrow$ & GCC@0.1$\uparrow$ & Cap$\uparrow$ & mIoU$\uparrow$ & GCC@0.1$\uparrow$ & Cap$\uparrow$ & mIoU$\uparrow$ & GCC@0.1$\uparrow$ & Cap$\uparrow$ & mIoU$\uparrow$ & GCC@0.1$\uparrow$ \\
\midrule
\multicolumn{13}{l}{\textit{Joint caption and mask generation}} \\
Sa2VA-8B & 4.40 & 31.14 & 3.55 & 3.99 & 17.28 & 3.43 & 7.53 & 35.08 & 6.70 & 4.57 & 20.36 & 2.90 \\
VideoGLaMM-3.8B & 7.88 & 22.13 & 2.86 & 3.89 & 23.19 & 1.82 & 8.67 & 24.31 & 3.69 & 4.02 & 20.75 & 2.90 \\
UniPixel-7B & 1.83 & 2.93 & 1.08 & 1.56 & 2.67 & 1.00 & 15.06 & 17.06 & 6.12 & 7.11 & 9.01 & 3.26 \\
\midrule
\multicolumn{13}{l}{\textit{Captioning with separate localization}} \\
Veason-R1-3B & 17.88 & 30.85 & 13.59 & 8.89 & 12.22 & 5.32 & 22.99 & 29.25 & 14.93 & 11.48 & 12.24 & 4.67 \\
InternVL3-8B + SAM2 & 27.22 & 24.33 & 11.47 & 18.10 & 18.06 & 9.54 & 31.11 & 27.12 & 15.34 & 22.03 & 17.23 & 7.78 \\
Qwen3-VL-8B + SAM2 & 27.18 & 42.32 & 21.42 & 18.25 & 26.91 & 13.14 & 33.28 & 42.94 & 27.93 & 19.79 & 24.80 & 12.41 \\
InternVL3.5-8B + SAM2 & 26.54 & 20.74 & 11.78 & 19.89 & 18.23 & 7.91 & 31.85 & 18.77 & 13.40 & 22.49 & 18.34 & 7.77 \\
Qwen3.5-9B + SAM2 & 39.60 & 48.34 & 30.09 & 30.64 & 33.99 & 20.39 & 36.56 & 46.67 & 28.79 & 26.41 & 32.40 & 19.84 \\
\bottomrule
\end{tabular}
\end{table*}

\subsection{Main Results}
\label{sec:main_results}

\paragraph{Good captions do not ensure correct grounding.}
Table~\ref{tab:instance_results} shows that similar caption quality can correspond to substantially different joint caption and grounding scores.
InternVL3+SAM2 and Qwen3-VL+SAM2 have nearly identical appearance Cap, yet their GCC@0.1 scores are 10.98 and 19.31, respectively.
Evaluating descriptions alone would treat these methods as comparable, without accounting for the visual support provided by their associated tracks.
Table~\ref{tab:main_results} shows that joint evaluation can also reverse the ordering of models.
Qwen3.5+SAM2 has higher Cap than Qwen3-VL+SAM2, but lower GCC@0.1 and GCC@avg.
Thus, caption quality alone does not establish how well a model grounds the participants it describes.
The anomaly ranking reversal also holds with a second caption judge and volume IoU 
(Suppl.~Table~\ref{tab:supp_judges} and Suppl.~Sec.~\ref{sec:supp_temporal_aggregation}).

\begin{figure*}[!t]
  \centering
  \includegraphics[width=0.95\textwidth]{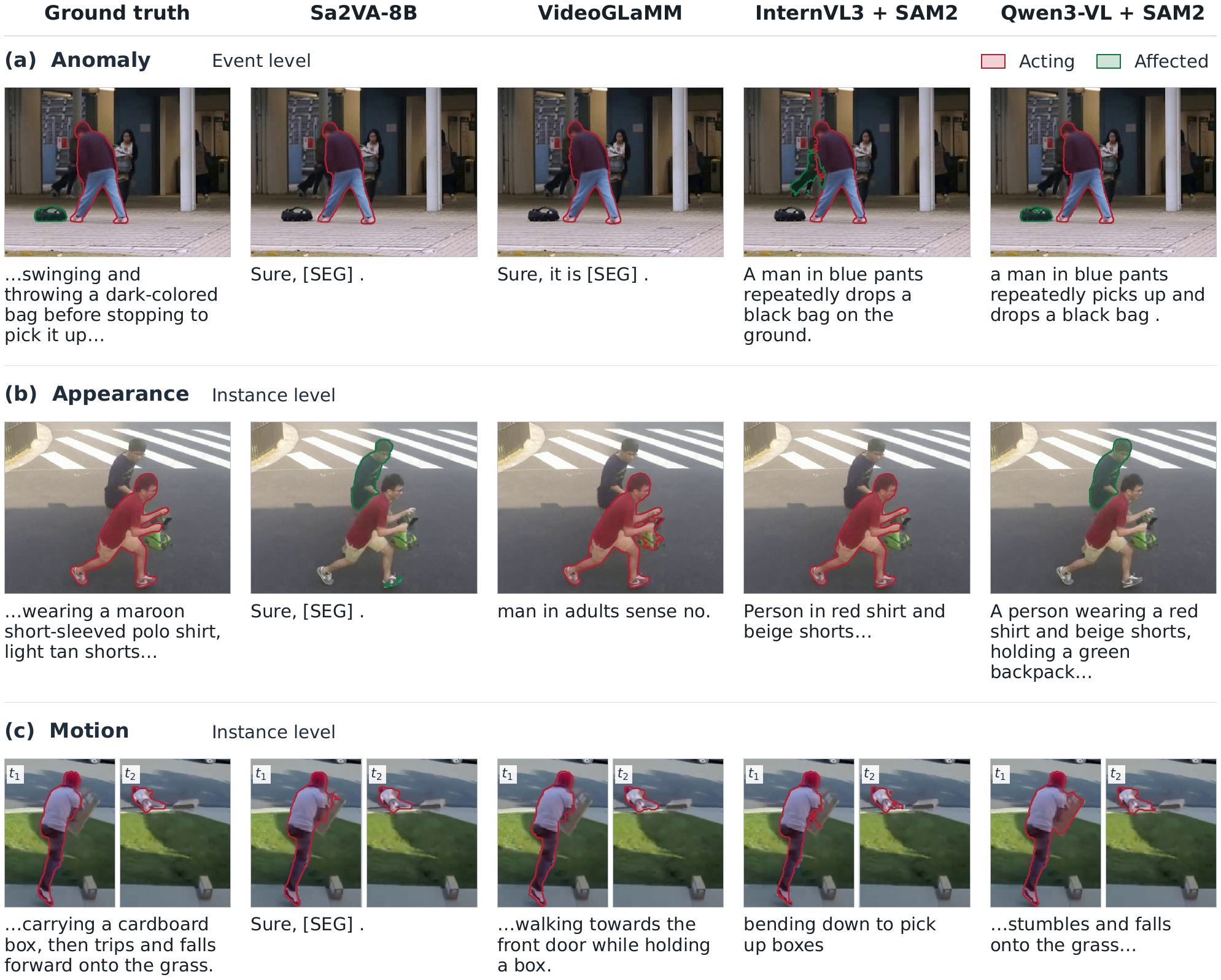}
  \caption{
  Qualitative comparison of \textbf{(a)}~anomaly, \textbf{(b)}~appearance, and \textbf{(c)}~motion grounding.
  Colors indicate matched ground-truth roles: red for acting and green for affected.
  Instance panels show selected caption--mask pairs.
  }
  \label{fig:qualitative_results}
\end{figure*}

\paragraph{Mask generation does not ensure informative descriptions.}
Table~\ref{tab:main_results} shows the opposite limitation in Sa2VA and VideoGLaMM.
Both obtain higher acting and affected mIoU than UniPixel, but much lower anomaly Cap.
Figure~\ref{fig:qualitative_results} (a) illustrates this contrast, as both models localize the acting person but return a segmentation response without describing the event.
These failures are consistent with reduced instruction following after grounding fine-tuning, which can lead models to overfit simple response patterns in the training data~\citep{wu2025f}.
Tables~\ref{tab:main_results} and~\ref{tab:instance_results} show that Veason-R1 and the VLM+SAM2 pipelines obtain higher Cap than the joint generation models across all three tasks.
Veason-R1 nevertheless retains only about half of its anomaly Cap under GCC@0.1.
These complementary failures show why an anomaly explanation needs both an informative description and the correct visual support.

\subsection{Participant Grounding}
\label{sec:participant_grounding}

\paragraph{Role-aware grounding remains difficult.}
Table~\ref{tab:main_results} shows a further gap when evaluation uses the predicted roles and checks every annotated participant.
Qwen3-VL+SAM2 obtains GCC@0.1 of 33.67, but rGCC@0.1 of 24.24.
The role-aware score requires visual support for each participant within its declared role, beyond matching the acting and affected groups as unordered regions.
Sa2VA, VideoGLaMM, and UniPixel receive zero rGCC because their outputs do not specify which masks correspond to the acting and affected roles.
Qwen3.5+SAM2 achieves the highest rGCC@0.1, while Qwen3-VL+SAM2 achieves the highest rGCC@avg.
Their order changes when stricter localization thresholds are included, showing that role-aware performance also depends on the precision of the supporting masks.

\paragraph{Affected participants receive lower joint scores.}
Table~\ref{tab:main_results} shows lower affected mIoU in all eight models.
Table~\ref{tab:role_results} shows lower affected Cap and GCC in both tasks for all eight models.
Affected mIoU is also lower in every comparison except VideoGLaMM appearance.
For Qwen3-VL+SAM2, affected motion GCC is less than half of the acting score.
The difficulty extends to both describing affected participants and providing their visual evidence.
An explanation can therefore remain incomplete even when the acting participant is identified.

Higher joint scores can also accompany additional predictions.
Table~\ref{tab:instance_results} shows that Qwen3.5+SAM2 has higher appearance and motion GCC than Qwen3-VL+SAM2, but more than three times as many unmatched tracks per clip in both tasks.
Reporting FPI alongside GCC distinguishes gains in grounded descriptions from the additional tracks produced with them.

\subsection{Qualitative Analysis}
\label{sec:qualitative}

Figure~\ref{fig:qualitative_results} illustrates the errors behind the quantitative comparisons.
In Figure~\ref{fig:qualitative_results} (a), the anomaly caption identifies an affected object that its associated masks can miss.
InternVL3+SAM2 describes the man dropping a bag but fails to localize the bag in the displayed frame.
Qwen3-VL+SAM2 grounds both the man and the bag.
Figure~\ref{fig:qualitative_results} (b) shows why a description and a mask can each appear reasonable while their pairing is incorrect.
Both VLM+SAM2 captions describe the man in red.
InternVL3+SAM2 masks that person, while Qwen3-VL+SAM2 masks the man in dark clothing.

Figure~\ref{fig:qualitative_results} (c) demonstrates the reverse problem, in which localization succeeds but the linked description misses the anomalous action.
VideoGLaMM and InternVL3+SAM2 track the man, but their linked captions miss his fall.
Qwen3-VL+SAM2 captures both the person and the fall.
Together, these cases distinguish participant omissions, 
incorrect caption--mask pairings, 
and missing action descriptions, 
each of which can be overlooked when captions or masks are evaluated alone.

\section{Limitations}
\label{sec:limitations}
rGCC checks each role group and its individual members, 
but does not verify which phrase in an anomaly caption refers to which member of the same role.
AnomGround provides complementary instance captions that link each participant's appearance and motion to its own mask track.
Evaluating these captions assesses individual caption--mask correspondence alongside the event and role information in anomaly captions.
rGCC can also award credit when the required participants are grounded but the model predicts a role absent from the annotation.
We evaluate this additional error with rGCC-strict (Suppl.~Table~\ref{tab:supp_rgcc}).

\section{Conclusion}
\label{sec:conclusion}
We introduced AnomGround to evaluate whether video anomaly explanations are grounded in the participants they describe.
Across six VAD datasets, its human-reviewed annotations connect anomaly captions to acting and affected groups and individual captions to participant mask tracks.
These complementary annotations support joint evaluation of event roles, participant localization, and individual caption--mask correspondence.
Our results show that strong captioning performance does not ensure correct participant grounding, with affected participants remaining particularly difficult.
Models can also localize a participant without producing an informative description of the anomalous event.
Checking every participant within its declared role also reveals incomplete grounding that group overlap can conceal.
AnomGround enables evaluation of both what an anomaly explanation describes and where its evidence appear.

\clearpage

\section{LLM Usage Statement}
OpenAI Codex assisted with editing author-produced text and refining author-led experimental analyses, and generated code that the authors reviewed and edited.
For dataset annotation, Gemini 3 Flash Preview generated drafts of the appearance, motion, and anomaly captions.
Trained annotators reviewed every caption against the corresponding video and mask tracks and corrected errors before inclusion in AnomGround.
As described in the evaluation protocol, \texttt{gpt-6-luna} and Mistral Small 3.2 served as caption judges.
The authors take full responsibility for all submitted content, including content produced with LLM or AI-tool assistance.

{
    \small
    \bibliographystyle{ieeenat_fullname}
    \bibliography{main}
}

\clearpage

\clearpage
\setcounter{page}{1}
\maketitlesupplementary

\section{Benchmark Construction Details}
\label{sec:supp_benchmark_construction}

\subsection{Dataset Statistics}
\label{sec:supp_dataset_statistics}

Table~\ref{tab:supp_annotation_counts} reports frame-level mask counts and caption counts by source dataset.
Of the benchmark's instance tracks, 74.6\% are acting and 25.4\% are affected.
Each annotated instance has one appearance caption and one motion caption, and each clip has one anomaly caption.
Acting instances appear in 99.8\% of clips and affected instances in 36.4\%, with both roles present in 36.2\%.
In 34.0\% of clips, at least one role contains multiple instances.
A further 22.6\% contain exactly one acting and one affected instance, accounting for the remaining multiple-instance clips in Table~\ref{tab:dataset_stats}.

\begin{table}[t]
\caption{
Benchmark annotation statistics across six VAD datasets. 
Acting and Affected report frame-level instance-mask counts, 
while Captions reports the number of annotated captions.
}
\label{tab:supp_annotation_counts}
\centering
\scriptsize
\setlength{\tabcolsep}{6pt}
\resizebox{\columnwidth}{!}{
\begin{tabular}{lccc}
\toprule
\textbf{Dataset} & \textbf{Acting} & \textbf{Affected} & \textbf{Captions} \\
\midrule
Ped1 & 6,929 & 0 & 158 \\
Ped2 & 2,718 & 0 & 50 \\
Avenue & 5,651 & 1,844 & 125 \\
ShanghaiTech & 28,277 & 2,221 & 442 \\
UCF-Crime & 80,494 & 46,700 & 693 \\
MSAD & 33,545 & 19,811 & 470 \\
\midrule
\textbf{Total} & 157,614 & 70,576 & 1,938 \\
\bottomrule
\end{tabular}
}
\end{table}

\subsection{Annotators and Review}
\label{sec:supp_annotator_labor}
Seven internal researchers trained in video anomaly detection and mask annotation performed the annotation and review.
They identified participants, assigned roles, and recorded event descriptions.
They inspected every frame of every mask track and checked every caption against the video with masks overlaid.
Errors in masks, captions, and their correspondence were corrected directly.

\subsection{Instance Annotation and Mask Correction}
\label{sec:supp_mask_annotation}

Figure~\ref{fig:mask_annotation_pipeline} summarizes the mask annotation process.
Each instance received a bounding box on a key frame, with positive and negative point prompts where needed.
Temporal metadata recorded the event extent, instance entry and exit, and frames in which an instance was absent or occluded.
SAM~2 propagated the prompted masks over the active temporal spans.
Frames outside those spans or marked as missing were assigned empty masks.

Following the initial annotation, incorrect masks received one to three additional rounds of box and point refinement.
When these refinements did not resolve the errors, annotators edited the masks directly at pixel level.
This correction procedure supported dense tracks in videos with complex scenes, low resolution, and limited visibility.

\subsection{Caption Generation Details}
\label{sec:supp_caption_gen}
\label{sec:caption_gen}

\begin{figure*}[t]
  \centering
  \includegraphics[width=\textwidth]{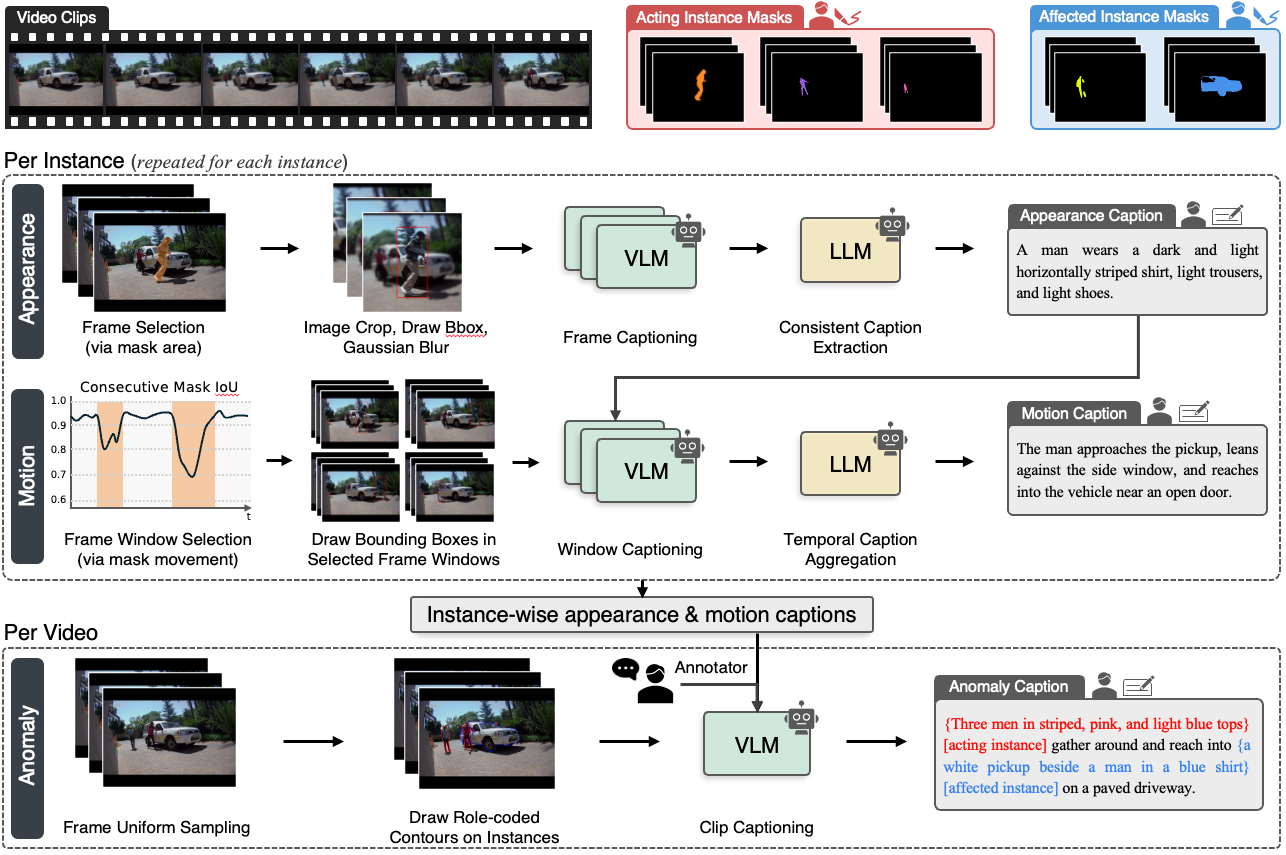}
  \caption{\textbf{Caption drafting and human review.}
  Reviewed mask tracks guide appearance and motion descriptions for each instance and an anomaly description for the event.
  Human icons indicate review and correction of every caption against the video.}
  \label{fig:supp_caption_pipeline}
\end{figure*}

Figure~\ref{fig:supp_caption_pipeline} presents the caption generation and review process using the inspected mask tracks.
We used Gemini 3 Flash Preview (\texttt{gemini-3-flash-preview}) to generate appearance, motion, and anomaly drafts and to consolidate the intermediate appearance and motion descriptions.
All generation calls used a temperature of 1.0, top-p of 1.0, and a maximum output length of 1,024 tokens.
Anomaly captions use role-coded mask overlays, instance captions, and the human-authored event description.
For clips with multiple annotated events, anomaly captions are generated separately and concatenated.
Annotators then review every caption alongside the mask-overlaid video and directly correct its content and its correspondence to the depicted instances.

\subsubsection{Appearance Caption}

\paragraph{Frame sampling.}
Frames are ranked by mask area in descending order.
We select up to 16 frames, initially keeping at least four frames between samples.
If fewer than 16 frames are selected, the remaining slots are filled in descending mask-area order without the spacing constraint.
Selected frames are presented in temporal order.

Each selected frame is cropped to a square region centered on the instance bounding box with a 20\% margin on each side (at least 30 pixels), clipped to image boundaries.
If the crop is smaller than $224 \times 224$ pixels, it is upscaled to at least 224 pixels on the longer side.
Regions outside an expanded bounding box are Gaussian-blurred with kernel size 11.

\paragraph{Per-image vision prompt.}
Given a cropped instance image with blurred background, the following prompt generates an appearance description.

\begin{quote}
\small\ttfamily
Describe the appearance of the highlighted object in detail.
- Focus on its visible attributes such as physical appearance, clothing colors, body type, distinctive features.
- The object can be a person, an animal, smoke, fire, vehicle, etc.
- DO NOT mention any bounding box, rectangle, or visual markers.
- The background and non-object regions have been intentionally blurred; ignore them completely and do not describe them.
- ONLY describe what the object looks like.
\end{quote}

Annotators provided appearance hints for instances that were difficult to recognize, typically objects other than people or vehicles.
We append the following text when a hint is available.

\begin{quote}
\small\ttfamily
Below is the brief human-provided hint of the appearance of this object.
Please use this hint as a reference only, and expand it into a detailed and comprehensive description.

human-provided appearance: [hint]
\end{quote}

\paragraph{Text synthesis prompt.}
Per-image descriptions are merged into a single appearance caption using a text-only synthesis prompt.

\begin{quote}
\small\ttfamily
You are given multiple appearance captions describing the same instance.
Synthesize them into one consistent and unified appearance caption.

Guidelines:
- Focus only on the object's physical appearance (e.g., physical appearance, clothing colors, body type, distinctive features).
- Ignore any information about actions, positions, or frame-specific details.
- If there are conflicting or inconsistent descriptions, ignore them and trust the attributes that appear repeatedly or consistently across the captions.
- If some details are mentioned only once and conflict with others, discard them.
- Ignore any references to red bounding boxes, rectangles, or visual annotations.
- Produce a single concise and coherent sentence.
\end{quote}

\subsubsection{Motion Caption}

\paragraph{Frame sampling.}
Frames with nonempty masks are ordered by time and divided into groups of up to $L\!=\!24$ visible frames.
Within each group, mask IoU values $\{u_j\}$ are computed between successive visible frames, with the first value set to zero.
A sliding window of size $W\!=\!6$ computes the sum $S_k = \sum_{j=k}^{k+W-1} u_j$ for each starting position $k$.
We select the window with the minimum sum $S^* = \min_k S_k$, using changes in mask overlap to select motion frames.
Groups with fewer than six frames use all available frames, while groups with fewer than two are skipped.
Selected frames are presented as full images with the instance highlighted by a bounding box with 10\% margin expansion.

\paragraph{Per-window vision prompt.}
Given a sequence of frames with the instance highlighted by a bounding box, the following prompt generates per-window motion descriptions.
The appearance caption is provided as context.

\begin{quote}
\small\ttfamily
You are given a sequence of video frames.
In each frame, the target object is highlighted with a red bounding box.

An appearance caption of the target object is provided below:
[appearance caption]

Your task is to describe the motion of the target object over time.

Focus only on:
- what the target object is doing over time (its actions or behaviors),
- how its state or position changes across frames as a result of these actions,
- interactions with other objects or the environment if relevant (without describing those objects in detail).

DO NOT:
- describe the appearance of the target object again,
- describe any static background details,
- describe objects other than the one in the red bounding box in detail.

Provide a concise and clear motion description in one or two sentences.
\end{quote}

\paragraph{Text synthesis prompt.}
Per-window descriptions are merged into a single motion caption.

\begin{quote}
\small\ttfamily
You are given motion descriptions from temporal segments:
[segment list]

Merge them into one final motion caption describing the target object's actions over the entire video.

Guidelines:
- Maintain the temporal order of actions.
- Remove redundancy.
- Summarize the overall behavior.
- Focus only on the target object's actions and interactions (without detailing other objects).

Output one concise paragraph (1--2 sentences).
\end{quote}

\subsubsection{Anomaly Caption}

\paragraph{Frame sampling.}
$N\!=\!8$ frames are sampled at uniform temporal intervals across the anomaly clip using indices $\lfloor i \cdot (T\!-\!1) / (N\!-\!1) \rfloor$ for $i = 0, \ldots, N\!-\!1$, where $T$ is the total number of frames.
Acting instance masks are rendered as red contours and affected instance masks as blue contours on the sampled frames.
When multiple instances share a role, all masks for that role are rendered in the same color.

\paragraph{Vision prompt.}
Given the uniformly sampled frames with acting (red contour) and affected (blue contour) instance overlays, per-instance structured descriptions, and the annotator-written event hint, the following prompt generates the role-grounded anomaly caption.

\begin{quote}
\small\ttfamily
You are given an image where anomalous objects are highlighted with colored contours.

Red contours indicate the anomaly acting instance (e.g., a person, fire, smoke, animal, or vehicle that triggers the abnormal event).
Blue contours indicate the anomaly affected instance (e.g., a person, object, or animal that is being affected or attacked).

For each highlighted object, you are provided with structured descriptions:
[per-instance appearance and motion captions]

Below is the brief human-provided hint of the anomalous event.
Please use this hint as a reference only, and expand it into a detailed and comprehensive description.
Each subject and object is represented using a role-and-index format (e.g., [acting instance 0], [affected instance 1]), where the index corresponds exactly to the instances defined in the structured descriptions above.
human-provided event:
[event text]

Your task is to generate a concise and natural language caption that describes the overall anomaly situation using both the image and the provided object-level descriptions.

Output Format:
- Enclose the anomaly-causing actor in curly braces and append [acting instance] ONLY ONCE in the entire caption.
- Enclose the anomaly target or victim in curly braces and append [affected instance] ONLY ONCE in the entire caption.
- If there is no acting or no affected instance in provided descriptions, DO NOT use the corresponding tag.

Example:
\{A speeding motorcycle and a red truck\} [acting instance] crashes into \{a pedestrian crossing the street\} [affected instance], causing a dangerous traffic accident.

Requirements:
1. Describe what abnormal or dangerous event is occurring.
2. Explain the interaction between the anomaly-acting instance (acting) and the target or victim (affected).
3. Integrate appearance and motion information when useful.
4. Do not mention contour colors, masks, bounding boxes, or annotations in the final caption.
5. Do not list instances separately; produce a single coherent event-level description.
6. Use clear action verbs (e.g., attacking, colliding, stealing, setting on fire, chasing, riding).
7. Limit the caption to one or two short sentences.
8. Even if there are multiple actors or multiple subjects, use the [acting instance] tag only once and the [affected instance] tag only once.
9. Do not repeat or duplicate the [acting instance] or [affected instance] tags.

Focus on answering:
What is happening? Who is causing it? Who is being affected?
\end{quote}

\subsection{Annotation Structure}
\label{sec:supp_schema}

Each clip records its events and the individual tracks assigned to each acting and affected group.
Each track retains its appearance and motion captions, entry and exit frames, and frames with no visible mask.
These temporal annotations determine which frames receive empty masks during evaluation.
Clips can contain multiple events with separate role assignments.
Individual tracks remain available alongside their role-group unions, supporting both collective event descriptions and individual caption--mask evaluation.

\FloatBarrier

\section{Evaluation and Baseline Details}
\label{sec:supp_eval_baseline}

Tables~\ref{tab:supp_event_uncertainty} and~\ref{tab:supp_instance_uncertainty} report the main results with population standard deviations across the three caption-judge rounds.

\subsection{Experimental Setup Details}
\label{sec:supp_exp_setup}

\paragraph{Baseline protocol.}
All baselines receive the same uniform sample of up to eight frames from each of the 412 dataset-labeled anomalous clips.
Clips shorter than eight frames use all available frames.
Masks are evaluated over every annotated frame of the anomaly clip.
No ground-truth role labels, instance IDs, or mask tracks are provided.
We prompt each baseline separately for appearance, motion, and anomaly outputs.
The VLM+SAM2 pipelines and caption-only models use a maximum response length of 1,024 new tokens.
For appearance and motion, the prompt asks the model to describe and, when possible, localize every anomaly-related instance.
For anomaly captions, every baseline receives the common prompt below, with native segmentation requests added where needed.
We remove role tags and formatting marks before scoring the anomaly caption.

\paragraph{Common anomaly prompt.}
The following prompt is shared across the anomaly-caption baselines.
Sa2VA and VideoGLaMM append a request for interleaved acting and affected segmentation masks.
UniPixel appends the segmentation request specified below.

\begin{quote}
\small\ttfamily
You are an expert vision-language model for anomaly understanding in video.

Task: the frames depict an anomalous situation. Describe the anomalous event in one complete sentence with a subject and a verb.

Rules:
- Wrap whatever causes the event in braces, followed by [acting instance].
- Wrap whatever it happens to in braces, followed by [affected instance].
- If nothing is affected, write no [affected instance].
- Write the verb outside the braces.
- Describe only these frames.

Two examples of the format:

\{a man in a green coat\} [acting instance] pushes \{a woman with a yellow bag\} [affected instance] onto a road.
\{a cyclist in a white helmet\} [acting instance] rides against the traffic on a bridge.
\end{quote}

\paragraph{Baseline selection.}
We evaluate models that can produce captions with mask tracks for the benchmark tasks.
VAD-specific systems such as HAWK~\citep{tang2024hawk} and Holmes-VAD~\citep{zhang2024holmes} provide event descriptions or answers and require additional localization components for this comparison.

\paragraph{Joint caption and mask generation.}
These models generate descriptions containing segmentation tokens that identify the corresponding mask tracks~\citep{rasheed2024glamm, munasinghe2025videoglamm}.
VideoGLaMM-3.8B~\citep{munasinghe2025videoglamm} generates text interleaved with segmentation tokens, which are decoded into per-frame masks through a spatiotemporal decoder.
Sa2VA~\citep{yuan2025sa2va} combines SAM~2 with a multimodal large language model.
Its generated segmentation tokens guide SAM~2 to produce instance masks.

We use the released Sa2VA-8B and VideoGLaMM checkpoints.
Their segmentation modules produce masks over the full clip.
VideoGLaMM uses its SAM~2 checkpoint and a fixed generation seed.
For appearance and motion, both models use the following request, with \texttt{[task]} replaced by the task name.

\begin{quote}
\small\ttfamily
Can you provide a thorough description of the [task] of anomaly related instances in the video?
Please respond with interleaved segmentation masks for the anomalous instances of the answer.
\end{quote}

UniPixel-7B~\citep{liu2026unipixel} generates a response and its mask tracks in one call using its official video input format and greedy decoding.
Each task question is followed by ``Please answer the question and provide the segmentation mask.''
For appearance and motion, we ask ``What is the appearance [or motion] of each person or object involved in the anomalous event in this video?'' and request one line per instance.
Each generated \texttt{<|seg|>} token produces a mask track through the model's SAM~2 decoder over the full clip.
For instance captions, each track is paired with the sentence containing its token, following the same pairing rule as Sa2VA.
We retain up to 15 tracks in response order.

\paragraph{Captioning with separate localization.}
These pipelines generate descriptions before localizing their participants, as in modular grounding systems~\citep{zhao2023bubogpt, munasinghe2023pgvideollava}.
The released Veason-R1-3B~\citep{gong2025reinforcing} uses two stages.
It first generates the requested caption from the eight input frames.
We extract participant phrases from that caption and pass each phrase to its video segmentation procedure.
Anomaly phrases come directly from the model's role tags.
For Veason-R1 appearance and motion captions, noun phrases are extracted with a deterministic dependency parser.
Veason-R1 predicts a keyframe and boxes that prompt SAM~2 tracking over the full anomaly clip.
When several phrases describe the same instance or event role, their masks are combined into one track.

In the VLM+SAM2 pipelines, each VLM first generates the requested descriptions from the eight input frames.
The box prediction calls continue that conversation, so they can use the generated descriptions.
SAM~2 propagates the boxes over the anomaly clip.
The four VLM backbones are InternVL3-8B, InternVL3.5-8B, Qwen3-VL-8B, and Qwen3.5-9B.
We disable reasoning mode for Qwen3.5.
For anomaly captions, the VLM predicts boxes for each role-tagged phrase and SAM~2 propagates them.
Tracks assigned to the same declared role are combined at each frame.
Appearance and motion outputs associate tracks with descriptions by output order.

\paragraph{Instance description prompt.}
The VLM+SAM2 backbones and Veason-R1 request instance descriptions in JavaScript Object Notation (JSON) with the following prompt.
The placeholder \texttt{[task]} is replaced by appearance or motion.

\begin{quote}
\small\ttfamily
You are an expert vision-language model for anomaly understanding in images.

Task: The given video frames depicts an anomalous situation.
Your goal is to detect ALL instances that are directly related to the anomaly and output their [task] description.

IMPORTANT:
- Focus ONLY on instances that are acting or affected by the anomalous event.
- Ignore background or irrelevant objects that are not related to the anomaly.

For each detected anomalous-related instance:
1. Provide a unique instance\_id.
2. Provide a description about [task] of this instance.

Output format (JSON only, no extra text):

\{"instances": [\{"instance\_id": 0,
"description": "[task] of this instance"\}, ...]\}
\end{quote}

\subsection{Additional Evaluation and Reporting Details}
\label{sec:supp_reporting}

Anomaly GCC treats predictions as an unordered set and measures localization of the ground-truth role groups.
rGCC evaluates each declared role group and every annotated member of that group.
If the model does not provide a usable role assignment for a required track, that event receives zero rGCC.
Role assignments come from explicit links between the model's declared roles and its predicted tracks, without inferring a role from track order.
For instance captions, the text-based assignment and the IoU assignment are computed independently.
GCC always uses the caption attached to the IoU-matched prediction.
J\&F averages region similarity (J, measured by IoU) and contour accuracy (F, measured by the boundary F-measure) over the annotated frames.
For appearance and motion, J\&F uses a one-to-one assignment that maximizes total track J\&F.
For anomaly captions, J\&F is evaluated on the IoU-matched role-group tracks.

\paragraph{Local instance evaluation.}
For role $r$ and frame $t$, let $P_{r,t}$ be the union of predicted masks declaring that role and $U_{r,t}$ the union of its ground-truth instance masks $G_{j,t}$.
We use the spatial partitioning principle of local segmentation evaluation~\citep{jaus2025every}, with annotated instance masks as the reference sites.
When at least one member is visible, each false-positive pixel outside $U_{r,t}$ is assigned to the nearest visible ground-truth instance of role $r$ by Euclidean distance to its mask.
All pixels in the image are included, and exact distance ties are divided equally among the tied instances.
Let $E_{j,t}$ be the resulting false-positive pixel count assigned to instance $j$.
For a visible instance, the local IoU is
\begin{equation}
  L_{j,t}=\frac{|P_{r(j),t}\cap G_{j,t}|}{|G_{j,t}|+E_{j,t}}.
  \label{eq:member_iou}
\end{equation}
The denominator includes missed pixels of instance $j$ and its assigned false positives, while excluding other members of the same role.
When ground-truth masks overlap, each member retains all of its own ground-truth pixels.
This preserves a score of one for a prediction equal to the full role union.
The spatial assignment depends only on the ground truth and does not infer or change the model's declared roles.

Let $\mathcal{T}_j$ contain frames with a nonempty $G_{j,t}$, and $\mathcal{A}_r$ contain frames with a nonempty $P_{r,t}\cup U_{r,t}$.
We compute the member and group scores as
\begin{equation}
  \begin{aligned}
  M_j&=\frac{1}{|\mathcal{T}_j|}\sum_{t\in\mathcal{T}_j}L_{j,t},\\
  H_r&=\frac{1}{|\mathcal{A}_r|}\sum_{t\in\mathcal{A}_r}
       \mathrm{IoU}(P_{r,t},U_{r,t}).
  \end{aligned}
  \label{eq:member_temporal}
\end{equation}
Each visible frame contributes equally to its member's score.
Jointly empty frames add no credit, while predictions made when the whole role group is absent reduce $H_r$.
For clip $i$, let $\mathcal{R}_i$ and $\mathcal{I}_i$ denote its annotated roles and individual instances.
The role-aware caption score is
\begin{equation}
  \begin{aligned}
  Q_i&=\min\!\left\{\min_{r\in\mathcal{R}_i}H_{i,r},
                   \min_{j\in\mathcal{I}_i}M_{i,j}\right\},\\
  \mathrm{rGCC}@\tau
  &=\frac{1}{N}\sum_{i=1}^{N}\mathrm{Cap}_i\,\mathbf{1}[Q_i\geq\tau].
  \end{aligned}
  \label{eq:rgcc}
\end{equation}
Missing role declarations or required mask predictions give zero rGCC.
For positive $\tau$, every member needs positive pixel intersection to pass the gate.
A completely missed member has $M_j=0$, so even a high group IoU cannot preserve caption credit.
This evaluates the spatial support of all annotated members without requiring separate predicted instance tracks.
It does not recover individual phrase-to-track identities within a group.
The complementary appearance and motion evaluations assess the correspondence between each instance caption and its associated track.

Let $\widehat{\mathcal{R}}_i$ be the set of roles declared in the predicted anomaly caption, including declared roles for which mask generation fails.
The stricter variant also evaluates whether the model predicts the correct set of roles:
\begin{equation}
  \mathrm{rGCC\text{-}strict}@\tau
  =\frac{1}{N}\sum_{i=1}^{N}\mathrm{Cap}_i\,
  \mathbf{1}[Q_i\geq\tau]\,
  \mathbf{1}[\widehat{\mathcal{R}}_i=\mathcal{R}_i].
  \label{eq:rgcc_strict}
\end{equation}
Thus, an extra affected role in an acting-only event removes caption credit even if the acting participants are localized.
This separates visual support for the annotated participants from predictions of roles absent from the annotation.
Table~\ref{tab:supp_rgcc} reports this additional role-set check.

\paragraph{Threshold averaging and reporting.}
GCC@avg is defined as
\begin{equation}
  \text{GCC@avg} = \frac{1}{9}\sum_{k=1}^{9}\mathrm{GCC}@\left(\frac{k}{10}\right).
  \label{eq:gcc_avg}
\end{equation}
rGCC@avg uses the same average with rGCC in place of GCC.
rGCC-strict@avg averages the strict score over the same nine thresholds, retaining caption credit only when the predicted and annotated role sets also agree.

For GCC and the separately reported mask metrics, track IoU averages frame IoU over every ground-truth frame of the anomaly clip.
A missing predicted frame is treated as an empty mask.
Following the DAVIS evaluation convention~\citep{Pont-Tuset_arXiv_2017, Caelles_arXiv_2019}, a frame scores one when both masks are empty, crediting a correct prediction of the instance's absence.
It scores zero when only one mask is empty.
GCC and recall additionally require positive pixel intersection with a matched prediction, so jointly empty frames alone do not earn grounding credit.
rGCC instead uses the frame sets in Eq.~\ref{eq:member_temporal} and requires every individual member to meet the threshold in Eq.~\ref{eq:rgcc}.
Missing or invalid mask predictions receive failure scores, while the saved text remains available for caption evaluation.

Recall is the fraction of ground-truth instances or role groups that satisfy the IoU threshold and have positive pixel intersection with a matched prediction.
We report false-positive instances (FPI) as the mean number of unmatched predicted tracks per clip.
For appearance and motion, a track is unmatched if it has no accepted match at IoU 0.1 with positive pixel intersection.
For anomaly, FPI counts tracks left unmatched by the one-to-one role-group assignment, which retains pairs with positive track IoU without applying the 0.1 threshold.
The count uses the tracks submitted for evaluation, after combining tracks with the same declared role for the role-tagged pipelines.

All aggregate metrics use pooled sums and denominators across the six datasets.
For appearance and motion, mIoU, J\&F, Cap, GCC, and recall are pooled over ground-truth instances.
Role-specific means pool the ground-truth instances with the corresponding acting or affected label.
For anomaly, Cap, GCC, and rGCC are pooled over video clips, while recall is pooled over annotated role groups.
Anomaly mIoU is averaged separately over all annotated role-group tracks of each role, as in Table~\ref{tab:main_results}.
FPI and logical VLM invocation counts are averaged over video clips.

Anomaly Cap, GCC, and rGCC are reported once per event description, without separate acting and affected caption scores.
Role-specific anomaly localization analyses report mIoU.
When an affected role is absent from a dataset, its spatial metric entries use a dash.
Caption-only methods also use a dash for spatial metrics.
Logical VLM invocation counts exclude segmentation, judge calls, and retries.
All scores except counts are reported on a 100-point scale, and GCC and rGCC use $\tau=0.1$ unless stated otherwise.
Mistral Small 3.2 at temperature zero provides a second LLM assessment.
We also report METEOR, a text similarity metric based on word alignment between the predicted and reference captions.

\subsection{LLM Judge Prompts}
\label{sec:supp_judge}

The $\mathrm{Cap}$ score is computed using \texttt{gpt-6-luna} with reasoning effort \texttt{none} as the LLM judge.
Each output level uses a separate prompt that focuses on the relevant semantic dimension.

\paragraph{Appearance judge prompt.}
\begin{quote}
\small\ttfamily
You are an expert evaluator for video anomaly detection captions.
Task: Compare a predicted instance appearance caption against a ground truth appearance caption and rate their similarity.

IMPORTANT: Evaluate ONLY the appearance of the instance (visual attributes such as clothing, color, shape, and physical characteristics).
Do NOT consider motion, actions, behaviors, or events.

Evaluation Criteria:
1. Visual Features: Do they describe the same appearance of the instance (clothing, objects, physical characteristics, colors, shape)?
2. Semantic Equivalence: Are the core meanings about appearance aligned, even if wording differs?

Rate the similarity on a scale of 0.0 to 1.0:
- 1.0: Perfect match (same appearance, semantically equivalent)
- 0.7--0.9: Good match (minor differences in detail or wording)
- 0.4--0.6: Partial match (some overlap but missing key appearance details)
- 0.1--0.3: Poor match (major discrepancies in appearance)
- 0.0: No match (completely different appearance)

Respond with ONLY a single float number between 0.0 and 1.0, nothing else.
\end{quote}

\paragraph{Motion judge prompt.}
\begin{quote}
\small\ttfamily
You are an expert evaluator for video anomaly detection captions.
Task: Compare a predicted instance motion caption against a ground truth motion caption and rate their similarity.

IMPORTANT: Evaluate ONLY the motion and actions of the instance (movement, behavior, interaction, and dynamics).
Do NOT consider appearance, clothing, color, shape, or physical attributes.

Evaluation Criteria:
1. Motion \& Actions: Do they describe the same movement, behavior, or interaction of the instance?
2. Semantic Equivalence: Are the core meanings about motion aligned, even if wording differs?

Rate the similarity on a scale of 0.0 to 1.0:
- 1.0: Perfect match (same motion/action, semantically equivalent)
- 0.7--0.9: Good match (minor differences in detail or wording)
- 0.4--0.6: Partial match (some overlap but missing key motion details)
- 0.1--0.3: Poor match (major discrepancies in motion/action)
- 0.0: No match (completely different motion/action)

Respond with ONLY a single float number between 0.0 and 1.0, nothing else.
\end{quote}

\paragraph{Anomaly judge prompt.}
\begin{quote}
\small\ttfamily
You are an expert evaluator for video anomaly detection captions.
Task: Compare a predicted anomaly caption against a ground truth anomaly caption and rate their similarity.

IMPORTANT: Evaluate whether both captions describe the SAME anomaly event.
Focus on the abnormal situation itself (what went wrong, who/what is involved, and how the anomaly occurs).
Consider event type, interaction, cause, and affected instances.
Do NOT penalize differences in wording or sentence structure if the meaning is the same.

Evaluation Criteria:
1. Anomaly Type: Do they describe the same kind of abnormal event (e.g., assault, accident, fire, fall, collision, intrusion)?
2. Participants \& Interaction: Are the involved instances (who/what causes and who/what is affected) consistent?
3. Semantic Equivalence: Are the core meanings of the anomaly aligned, even if wording differs?

Rate the similarity on a scale of 0.0 to 1.0:
- 1.0: Perfect match (same anomaly event, semantically equivalent)
- 0.7--0.9: Good match (minor differences in detail or wording)
- 0.4--0.6: Partial match (some overlap but missing key anomaly aspects)
- 0.1--0.3: Poor match (major discrepancies in anomaly description)
- 0.0: No match (completely different anomaly event)

Respond with ONLY a single float number between 0.0 and 1.0, nothing else.
\end{quote}

\FloatBarrier

\section{Additional Results}
\label{sec:supp_additional_results}

\FloatBarrier

\subsection{Aggregate Results and Judge Variation}
\label{sec:supp_judge_variation}

Tables~\ref{tab:supp_event_uncertainty} and~\ref{tab:supp_instance_uncertainty} report the aggregate anomaly and instance results with judge variation.
Caption and joint scores are shown as the mean and population standard deviation across three independent judge rounds.
Caption and joint-score means match Tables~\ref{tab:main_results} and~\ref{tab:instance_results}.
The anomaly mIoU columns use the same acting and affected role-group tracks as Table~\ref{tab:main_results}.
Role-specific results with standard deviations are reported in Sec.~\ref{sec:supp_roles}.

UniPixel emits no segmentation token in 88.1\% of appearance responses, 46.6\% of motion responses, and 63.6\% of anomaly responses.
These responses provide no mask track, even when their text describes the video.
Its outputs also lack usable mask-to-role assignments, yielding zero rGCC.

\begin{table*}[t]
\centering
\caption{Anomaly captioning and grounding with judge standard deviations.}
\label{tab:supp_event_uncertainty}
\small
\setlength{\tabcolsep}{3pt}
\renewcommand{\arraystretch}{1.1}
\begin{tabular}{lcccccccc}
\toprule
\multirow{2}{*}{Model} & \multirow{2}{*}{Cap$\uparrow$} & \multicolumn{2}{c}{mIoU$\uparrow$} & \multicolumn{2}{c}{GCC$\uparrow$} & \multicolumn{2}{c}{rGCC$\uparrow$} & \multirow{2}{*}{FPI$\downarrow$} \\
\cmidrule(lr){3-4} \cmidrule(lr){5-6} \cmidrule(lr){7-8}
 & & Acting & Affected & @0.1 & @avg & @0.1 & @avg & \\
\midrule
\multicolumn{9}{l}{\textit{Joint caption and mask generation}} \\
Sa2VA-8B & \begin{tabular}[c]{@{}c@{}}$4.16$\\[-3pt]$\scriptstyle\pm 0.07$\end{tabular} & 36.35 & 12.34 & \begin{tabular}[c]{@{}c@{}}$0.19$\\[-3pt]$\scriptstyle\pm 0.01$\end{tabular} & \begin{tabular}[c]{@{}c@{}}$0.10$\\[-3pt]$\scriptstyle\pm 0.01$\end{tabular} & \begin{tabular}[c]{@{}c@{}}$0.00$\\[-3pt]$\scriptstyle\pm 0.00$\end{tabular} & \begin{tabular}[c]{@{}c@{}}$0.00$\\[-3pt]$\scriptstyle\pm 0.00$\end{tabular} & 0.107 \\
VideoGLaMM-3.8B & \begin{tabular}[c]{@{}c@{}}$2.35$\\[-3pt]$\scriptstyle\pm 0.04$\end{tabular} & 26.03 & 16.77 & \begin{tabular}[c]{@{}c@{}}$0.96$\\[-3pt]$\scriptstyle\pm 0.04$\end{tabular} & \begin{tabular}[c]{@{}c@{}}$0.68$\\[-3pt]$\scriptstyle\pm 0.03$\end{tabular} & \begin{tabular}[c]{@{}c@{}}$0.00$\\[-3pt]$\scriptstyle\pm 0.00$\end{tabular} & \begin{tabular}[c]{@{}c@{}}$0.00$\\[-3pt]$\scriptstyle\pm 0.00$\end{tabular} & 0.391 \\
UniPixel-7B & \begin{tabular}[c]{@{}c@{}}$20.23$\\[-3pt]$\scriptstyle\pm 0.18$\end{tabular} & 10.12 & 3.09 & \begin{tabular}[c]{@{}c@{}}$4.06$\\[-3pt]$\scriptstyle\pm 0.05$\end{tabular} & \begin{tabular}[c]{@{}c@{}}$2.79$\\[-3pt]$\scriptstyle\pm 0.04$\end{tabular} & \begin{tabular}[c]{@{}c@{}}$0.00$\\[-3pt]$\scriptstyle\pm 0.00$\end{tabular} & \begin{tabular}[c]{@{}c@{}}$0.00$\\[-3pt]$\scriptstyle\pm 0.00$\end{tabular} & 0.100 \\
\midrule
\multicolumn{9}{l}{\textit{Captioning with separate localization}} \\
Veason-R1-3B & \begin{tabular}[c]{@{}c@{}}$25.20$\\[-3pt]$\scriptstyle\pm 0.15$\end{tabular} & 43.09 & 33.59 & \begin{tabular}[c]{@{}c@{}}$13.96$\\[-3pt]$\scriptstyle\pm 0.05$\end{tabular} & \begin{tabular}[c]{@{}c@{}}$9.50$\\[-3pt]$\scriptstyle\pm 0.04$\end{tabular} & \begin{tabular}[c]{@{}c@{}}$8.61$\\[-3pt]$\scriptstyle\pm 0.02$\end{tabular} & \begin{tabular}[c]{@{}c@{}}$6.36$\\[-3pt]$\scriptstyle\pm 0.01$\end{tabular} & 0.898 \\
InternVL3-8B + SAM2 & \begin{tabular}[c]{@{}c@{}}$40.87$\\[-3pt]$\scriptstyle\pm 0.12$\end{tabular} & 27.35 & 25.78 & \begin{tabular}[c]{@{}c@{}}$19.54$\\[-3pt]$\scriptstyle\pm 0.12$\end{tabular} & \begin{tabular}[c]{@{}c@{}}$8.12$\\[-3pt]$\scriptstyle\pm 0.05$\end{tabular} & \begin{tabular}[c]{@{}c@{}}$12.15$\\[-3pt]$\scriptstyle\pm 0.11$\end{tabular} & \begin{tabular}[c]{@{}c@{}}$5.56$\\[-3pt]$\scriptstyle\pm 0.04$\end{tabular} & 0.575 \\
Qwen3-VL-8B + SAM2 & \begin{tabular}[c]{@{}c@{}}$43.98$\\[-3pt]$\scriptstyle\pm 0.16$\end{tabular} & 50.88 & 38.40 & \begin{tabular}[c]{@{}c@{}}$33.67$\\[-3pt]$\scriptstyle\pm 0.12$\end{tabular} & \begin{tabular}[c]{@{}c@{}}$20.85$\\[-3pt]$\scriptstyle\pm 0.08$\end{tabular} & \begin{tabular}[c]{@{}c@{}}$24.24$\\[-3pt]$\scriptstyle\pm 0.11$\end{tabular} & \begin{tabular}[c]{@{}c@{}}$16.02$\\[-3pt]$\scriptstyle\pm 0.10$\end{tabular} & 0.464 \\
InternVL3.5-8B + SAM2 & \begin{tabular}[c]{@{}c@{}}$37.26$\\[-3pt]$\scriptstyle\pm 0.12$\end{tabular} & 16.34 & 14.93 & \begin{tabular}[c]{@{}c@{}}$14.31$\\[-3pt]$\scriptstyle\pm 0.04$\end{tabular} & \begin{tabular}[c]{@{}c@{}}$3.88$\\[-3pt]$\scriptstyle\pm 0.01$\end{tabular} & \begin{tabular}[c]{@{}c@{}}$6.76$\\[-3pt]$\scriptstyle\pm 0.05$\end{tabular} & \begin{tabular}[c]{@{}c@{}}$2.01$\\[-3pt]$\scriptstyle\pm 0.01$\end{tabular} & 0.502 \\
Qwen3.5-9B + SAM2 & \begin{tabular}[c]{@{}c@{}}$45.54$\\[-3pt]$\scriptstyle\pm 0.36$\end{tabular} & 47.49 & 35.86 & \begin{tabular}[c]{@{}c@{}}$33.21$\\[-3pt]$\scriptstyle\pm 0.24$\end{tabular} & \begin{tabular}[c]{@{}c@{}}$18.55$\\[-3pt]$\scriptstyle\pm 0.16$\end{tabular} & \begin{tabular}[c]{@{}c@{}}$24.68$\\[-3pt]$\scriptstyle\pm 0.12$\end{tabular} & \begin{tabular}[c]{@{}c@{}}$13.70$\\[-3pt]$\scriptstyle\pm 0.12$\end{tabular} & 0.510 \\
\bottomrule
\end{tabular}
\end{table*}

\begin{table*}[t]
\centering
\caption{Instance scores with judge standard deviations.}
\label{tab:supp_instance_uncertainty}
\small
\setlength{\tabcolsep}{2pt}
\renewcommand{\arraystretch}{1.1}
\begin{tabular}{lcccccccccc}
\toprule
Model & \multicolumn{5}{c}{\textbf{Appearance}} & \multicolumn{5}{c}{\textbf{Motion}} \\
\cmidrule(lr){2-6} \cmidrule(lr){7-11}
 & Cap$\uparrow$ & mIoU$\uparrow$ & GCC@0.1$\uparrow$ & GCC@avg$\uparrow$ & FPI$\downarrow$ & Cap$\uparrow$ & mIoU$\uparrow$ & GCC@0.1$\uparrow$ & GCC@avg$\uparrow$ & FPI$\downarrow$ \\
\midrule
\multicolumn{11}{l}{\textit{Joint caption and mask generation}} \\
Sa2VA-8B & \begin{tabular}[c]{@{}c@{}}$4.30$\\[-3pt]$\scriptstyle\pm 0.02$\end{tabular} & 27.62 & \begin{tabular}[c]{@{}c@{}}$3.52$\\[-3pt]$\scriptstyle\pm 0.03$\end{tabular} & \begin{tabular}[c]{@{}c@{}}$2.43$\\[-3pt]$\scriptstyle\pm 0.02$\end{tabular} & 0.449 & \begin{tabular}[c]{@{}c@{}}$6.78$\\[-3pt]$\scriptstyle\pm 0.05$\end{tabular} & 31.34 & \begin{tabular}[c]{@{}c@{}}$5.74$\\[-3pt]$\scriptstyle\pm 0.02$\end{tabular} & \begin{tabular}[c]{@{}c@{}}$3.69$\\[-3pt]$\scriptstyle\pm 0.01$\end{tabular} & 0.806 \\
VideoGLaMM-3.8B & \begin{tabular}[c]{@{}c@{}}$6.87$\\[-3pt]$\scriptstyle\pm 0.04$\end{tabular} & 22.40 & \begin{tabular}[c]{@{}c@{}}$2.60$\\[-3pt]$\scriptstyle\pm 0.01$\end{tabular} & \begin{tabular}[c]{@{}c@{}}$1.87$\\[-3pt]$\scriptstyle\pm 0.01$\end{tabular} & 1.223 & \begin{tabular}[c]{@{}c@{}}$7.49$\\[-3pt]$\scriptstyle\pm 0.07$\end{tabular} & 23.41 & \begin{tabular}[c]{@{}c@{}}$3.49$\\[-3pt]$\scriptstyle\pm 0.04$\end{tabular} & \begin{tabular}[c]{@{}c@{}}$2.47$\\[-3pt]$\scriptstyle\pm 0.03$\end{tabular} & 1.148 \\
UniPixel-7B & \begin{tabular}[c]{@{}c@{}}$1.76$\\[-3pt]$\scriptstyle\pm 0.01$\end{tabular} & 2.87 & \begin{tabular}[c]{@{}c@{}}$1.06$\\[-3pt]$\scriptstyle\pm 0.01$\end{tabular} & \begin{tabular}[c]{@{}c@{}}$0.79$\\[-3pt]$\scriptstyle\pm 0.01$\end{tabular} & 0.097 & \begin{tabular}[c]{@{}c@{}}$13.04$\\[-3pt]$\scriptstyle\pm 0.04$\end{tabular} & 15.01 & \begin{tabular}[c]{@{}c@{}}$5.39$\\[-3pt]$\scriptstyle\pm 0.03$\end{tabular} & \begin{tabular}[c]{@{}c@{}}$3.64$\\[-3pt]$\scriptstyle\pm 0.02$\end{tabular} & 1.036 \\
\midrule
\multicolumn{11}{l}{\textit{Captioning with separate localization}} \\
Veason-R1-3B & \begin{tabular}[c]{@{}c@{}}$15.60$\\[-3pt]$\scriptstyle\pm 0.10$\end{tabular} & 26.12 & \begin{tabular}[c]{@{}c@{}}$11.49$\\[-3pt]$\scriptstyle\pm 0.06$\end{tabular} & \begin{tabular}[c]{@{}c@{}}$9.41$\\[-3pt]$\scriptstyle\pm 0.05$\end{tabular} & 0.461 & \begin{tabular}[c]{@{}c@{}}$20.07$\\[-3pt]$\scriptstyle\pm 0.06$\end{tabular} & 24.93 & \begin{tabular}[c]{@{}c@{}}$12.32$\\[-3pt]$\scriptstyle\pm 0.03$\end{tabular} & \begin{tabular}[c]{@{}c@{}}$8.82$\\[-3pt]$\scriptstyle\pm 0.02$\end{tabular} & 0.638 \\
InternVL3-8B + SAM2 & \begin{tabular}[c]{@{}c@{}}$24.90$\\[-3pt]$\scriptstyle\pm 0.01$\end{tabular} & 22.74 & \begin{tabular}[c]{@{}c@{}}$10.98$\\[-3pt]$\scriptstyle\pm 0.07$\end{tabular} & \begin{tabular}[c]{@{}c@{}}$5.30$\\[-3pt]$\scriptstyle\pm 0.03$\end{tabular} & 1.032 & \begin{tabular}[c]{@{}c@{}}$28.80$\\[-3pt]$\scriptstyle\pm 0.04$\end{tabular} & 24.61 & \begin{tabular}[c]{@{}c@{}}$13.42$\\[-3pt]$\scriptstyle\pm 0.09$\end{tabular} & \begin{tabular}[c]{@{}c@{}}$6.20$\\[-3pt]$\scriptstyle\pm 0.04$\end{tabular} & 1.199 \\
Qwen3-VL-8B + SAM2 & \begin{tabular}[c]{@{}c@{}}$24.91$\\[-3pt]$\scriptstyle\pm 0.22$\end{tabular} & 38.40 & \begin{tabular}[c]{@{}c@{}}$19.31$\\[-3pt]$\scriptstyle\pm 0.17$\end{tabular} & \begin{tabular}[c]{@{}c@{}}$14.60$\\[-3pt]$\scriptstyle\pm 0.13$\end{tabular} & 0.425 & \begin{tabular}[c]{@{}c@{}}$29.85$\\[-3pt]$\scriptstyle\pm 0.04$\end{tabular} & 38.33 & \begin{tabular}[c]{@{}c@{}}$23.98$\\[-3pt]$\scriptstyle\pm 0.03$\end{tabular} & \begin{tabular}[c]{@{}c@{}}$17.20$\\[-3pt]$\scriptstyle\pm 0.03$\end{tabular} & 0.444 \\
InternVL3.5-8B + SAM2 & \begin{tabular}[c]{@{}c@{}}$24.85$\\[-3pt]$\scriptstyle\pm 0.14$\end{tabular} & 20.10 & \begin{tabular}[c]{@{}c@{}}$10.80$\\[-3pt]$\scriptstyle\pm 0.09$\end{tabular} & \begin{tabular}[c]{@{}c@{}}$3.41$\\[-3pt]$\scriptstyle\pm 0.04$\end{tabular} & 1.400 & \begin{tabular}[c]{@{}c@{}}$29.47$\\[-3pt]$\scriptstyle\pm 0.12$\end{tabular} & 18.66 & \begin{tabular}[c]{@{}c@{}}$11.97$\\[-3pt]$\scriptstyle\pm 0.04$\end{tabular} & \begin{tabular}[c]{@{}c@{}}$3.87$\\[-3pt]$\scriptstyle\pm 0.01$\end{tabular} & 1.573 \\
Qwen3.5-9B + SAM2 & \begin{tabular}[c]{@{}c@{}}$37.32$\\[-3pt]$\scriptstyle\pm 0.03$\end{tabular} & 44.69 & \begin{tabular}[c]{@{}c@{}}$27.62$\\[-3pt]$\scriptstyle\pm 0.02$\end{tabular} & \begin{tabular}[c]{@{}c@{}}$19.46$\\[-3pt]$\scriptstyle\pm 0.01$\end{tabular} & 1.403 & \begin{tabular}[c]{@{}c@{}}$33.98$\\[-3pt]$\scriptstyle\pm 0.02$\end{tabular} & 43.04 & \begin{tabular}[c]{@{}c@{}}$26.51$\\[-3pt]$\scriptstyle\pm 0.03$\end{tabular} & \begin{tabular}[c]{@{}c@{}}$17.47$\\[-3pt]$\scriptstyle\pm 0.04$\end{tabular} & 1.515 \\
\bottomrule
\end{tabular}
\end{table*}

\FloatBarrier

\subsection{Results by Participant Role}
\label{sec:supp_roles}

Tables~\ref{tab:supp_role_appearance} and~\ref{tab:supp_role_motion} report appearance and motion results separately for acting and affected instances.
Each score pools the corresponding numerators and denominators over all annotated instances of that role across the six datasets.
The role breakdown distinguishes description quality, localization, and their correspondence when comparing acting and affected participants.

\begin{table*}[t]
\centering
\caption{Appearance captioning by participant role.}
\label{tab:supp_role_appearance}
\footnotesize
\setlength{\tabcolsep}{3pt}
\renewcommand{\arraystretch}{1.1}
\begin{tabular}{lcccccccc}
\toprule
Model & \multicolumn{4}{c}{Acting} & \multicolumn{4}{c}{Affected} \\
\cmidrule(lr){2-5} \cmidrule(lr){6-9}
 & mIoU & J\&F & Cap & GCC@0.1 & mIoU & J\&F & Cap & GCC@0.1 \\
\midrule
\multicolumn{9}{l}{\textit{Joint caption and mask generation}} \\
Sa2VA-8B & 31.14 & 33.55 & $4.40_{\pm 0.05}$ & $3.55_{\pm 0.04}$ & 17.28 & 18.78 & $3.99_{\pm 0.06}$ & $3.43_{\pm 0.04}$ \\
VideoGLaMM-3.8B & 22.13 & 23.49 & $7.88_{\pm 0.03}$ & $2.86_{\pm 0.02}$ & 23.19 & 24.35 & $3.89_{\pm 0.18}$ & $1.82_{\pm 0.10}$ \\
UniPixel-7B & 2.93 & 3.16 & $1.83_{\pm 0.02}$ & $1.08_{\pm 0.01}$ & 2.67 & 2.80 & $1.56_{\pm 0.06}$ & $1.00_{\pm 0.02}$ \\
\midrule
\multicolumn{9}{l}{\textit{Captioning with separate localization}} \\
Veason-R1-3B & 30.85 & 32.36 & $17.88_{\pm 0.04}$ & $13.59_{\pm 0.08}$ & 12.22 & 13.00 & $8.89_{\pm 0.26}$ & $5.32_{\pm 0.02}$ \\
InternVL3-8B + SAM2 & 24.33 & 25.89 & $27.22_{\pm 0.02}$ & $11.47_{\pm 0.13}$ & 18.06 & 19.57 & $18.10_{\pm 0.03}$ & $9.54_{\pm 0.15}$ \\
Qwen3-VL-8B + SAM2 & 42.32 & 44.55 & $27.18_{\pm 0.19}$ & $21.42_{\pm 0.18}$ & 26.91 & 28.40 & $18.25_{\pm 0.29}$ & $13.14_{\pm 0.14}$ \\
InternVL3.5-8B + SAM2 & 20.74 & 23.75 & $26.54_{\pm 0.18}$ & $11.78_{\pm 0.13}$ & 18.23 & 20.00 & $19.89_{\pm 0.23}$ & $7.91_{\pm 0.04}$ \\
Qwen3.5-9B + SAM2 & 48.34 & 50.97 & $39.60_{\pm 0.03}$ & $30.09_{\pm 0.03}$ & 33.99 & 36.12 & $30.64_{\pm 0.18}$ & $20.39_{\pm 0.04}$ \\
\bottomrule
\end{tabular}
\end{table*}

\begin{table*}[t]
\centering
\caption{Motion captioning by participant role.}
\label{tab:supp_role_motion}
\footnotesize
\setlength{\tabcolsep}{3pt}
\renewcommand{\arraystretch}{1.1}
\begin{tabular}{lcccccccc}
\toprule
Model & \multicolumn{4}{c}{Acting} & \multicolumn{4}{c}{Affected} \\
\cmidrule(lr){2-5} \cmidrule(lr){6-9}
 & mIoU & J\&F & Cap & GCC@0.1 & mIoU & J\&F & Cap & GCC@0.1 \\
\midrule
\multicolumn{9}{l}{\textit{Joint caption and mask generation}} \\
Sa2VA-8B & 35.08 & 37.79 & $7.53_{\pm 0.02}$ & $6.70_{\pm 0.03}$ & 20.36 & 22.21 & $4.57_{\pm 0.17}$ & $2.90_{\pm 0.02}$ \\
VideoGLaMM-3.8B & 24.31 & 25.90 & $8.67_{\pm 0.17}$ & $3.69_{\pm 0.06}$ & 20.75 & 21.88 & $4.02_{\pm 0.40}$ & $2.90_{\pm 0.06}$ \\
UniPixel-7B & 17.06 & 18.14 & $15.06_{\pm 0.04}$ & $6.12_{\pm 0.02}$ & 9.01 & 10.08 & $7.11_{\pm 0.18}$ & $3.26_{\pm 0.06}$ \\
\midrule
\multicolumn{9}{l}{\textit{Captioning with separate localization}} \\
Veason-R1-3B & 29.25 & 30.98 & $22.99_{\pm 0.05}$ & $14.93_{\pm 0.04}$ & 12.24 & 13.51 & $11.48_{\pm 0.25}$ & $4.67_{\pm 0.02}$ \\
InternVL3-8B + SAM2 & 27.12 & 29.03 & $31.11_{\pm 0.07}$ & $15.34_{\pm 0.10}$ & 17.23 & 18.32 & $22.03_{\pm 0.36}$ & $7.78_{\pm 0.04}$ \\
Qwen3-VL-8B + SAM2 & 42.94 & 45.20 & $33.28_{\pm 0.04}$ & $27.93_{\pm 0.02}$ & 24.80 & 26.50 & $19.79_{\pm 0.06}$ & $12.41_{\pm 0.14}$ \\
InternVL3.5-8B + SAM2 & 18.77 & 21.40 & $31.85_{\pm 0.25}$ & $13.40_{\pm 0.02}$ & 18.34 & 19.73 & $22.49_{\pm 0.25}$ & $7.77_{\pm 0.16}$ \\
Qwen3.5-9B + SAM2 & 46.67 & 49.02 & $36.56_{\pm 0.12}$ & $28.79_{\pm 0.08}$ & 32.40 & 34.38 & $26.41_{\pm 0.30}$ & $19.84_{\pm 0.13}$ \\
\bottomrule
\end{tabular}
\end{table*}

\FloatBarrier

\subsection{Results by Dataset}
\label{sec:supp_dataset_results}

Tables~\ref{tab:supp_dataset_anomaly}--\ref{tab:supp_dataset_motion} report GCC@0.1 for each source dataset.
All values use the same eight-frame input and full-clip mask evaluation as the main results.
Scores are multiplied by 100 and show the mean and population standard deviation over three independent judge rounds.

\begin{table*}[t]
\centering
\caption{Anomaly GCC@0.1 across source datasets.}
\label{tab:supp_dataset_anomaly}
\footnotesize
\setlength{\tabcolsep}{3pt}
\renewcommand{\arraystretch}{1.1}
\begin{tabular}{lcccccc}
\toprule
Model & Ped1 & Ped2 & Avenue & ShanghaiTech & UCF-Crime & MSAD \\
\midrule
\multicolumn{7}{l}{\textit{Joint caption and mask generation}} \\
Sa2VA-8B & $0.00_{\pm 0.00}$ & $0.00_{\pm 0.00}$ & $0.00_{\pm 0.00}$ & $0.00_{\pm 0.00}$ & $0.51_{\pm 0.04}$ & $0.11_{\pm 0.00}$ \\
VideoGLaMM-3.8B & $0.00_{\pm 0.00}$ & $0.00_{\pm 0.00}$ & $0.22_{\pm 0.15}$ & $0.19_{\pm 0.00}$ & $1.48_{\pm 0.07}$ & $1.88_{\pm 0.10}$ \\
UniPixel-7B & $12.17_{\pm 0.42}$ & $0.00_{\pm 0.00}$ & $6.13_{\pm 0.00}$ & $5.29_{\pm 0.00}$ & $2.52_{\pm 0.06}$ & $1.24_{\pm 0.05}$ \\
\midrule
\multicolumn{7}{l}{\textit{Captioning with separate localization}} \\
Veason-R1-3B & $14.33_{\pm 0.12}$ & $12.22_{\pm 0.39}$ & $30.38_{\pm 0.33}$ & $18.37_{\pm 0.24}$ & $10.59_{\pm 0.24}$ & $8.44_{\pm 0.13}$ \\
InternVL3-8B + SAM2 & $26.50_{\pm 0.00}$ & $31.53_{\pm 0.20}$ & $24.35_{\pm 0.35}$ & $23.62_{\pm 0.33}$ & $12.82_{\pm 0.24}$ & $18.32_{\pm 0.15}$ \\
Qwen3-VL-8B + SAM2 & $41.83_{\pm 0.31}$ & $49.44_{\pm 0.39}$ & $46.29_{\pm 0.47}$ & $36.79_{\pm 0.47}$ & $27.49_{\pm 0.16}$ & $29.17_{\pm 0.05}$ \\
InternVL3.5-8B + SAM2 & $22.38_{\pm 0.51}$ & $33.06_{\pm 0.39}$ & $16.83_{\pm 0.27}$ & $19.18_{\pm 0.07}$ & $8.40_{\pm 0.11}$ & $10.51_{\pm 0.14}$ \\
Qwen3.5-9B + SAM2 & $45.29_{\pm 0.21}$ & $39.72_{\pm 0.39}$ & $41.08_{\pm 0.15}$ & $27.77_{\pm 0.33}$ & $32.32_{\pm 0.24}$ & $31.90_{\pm 0.21}$ \\
\bottomrule
\end{tabular}
\end{table*}

\begin{table*}[t]
\centering
\caption{Appearance GCC@0.1 across source datasets.}
\label{tab:supp_dataset_appearance}
\footnotesize
\setlength{\tabcolsep}{3pt}
\renewcommand{\arraystretch}{1.1}
\begin{tabular}{lcccccc}
\toprule
Model & Ped1 & Ped2 & Avenue & ShanghaiTech & UCF-Crime & MSAD \\
\midrule
\multicolumn{7}{l}{\textit{Joint caption and mask generation}} \\
Sa2VA-8B & $8.05_{\pm 0.07}$ & $3.33_{\pm 0.25}$ & $0.00_{\pm 0.00}$ & $0.00_{\pm 0.00}$ & $7.63_{\pm 0.08}$ & $0.00_{\pm 0.00}$ \\
VideoGLaMM-3.8B & $1.19_{\pm 0.00}$ & $3.33_{\pm 0.25}$ & $3.19_{\pm 0.17}$ & $2.86_{\pm 0.03}$ & $2.30_{\pm 0.02}$ & $3.01_{\pm 0.03}$ \\
UniPixel-7B & $0.45_{\pm 0.08}$ & $1.05_{\pm 0.00}$ & $2.91_{\pm 0.10}$ & $0.65_{\pm 0.00}$ & $0.63_{\pm 0.02}$ & $1.79_{\pm 0.03}$ \\
\midrule
\multicolumn{7}{l}{\textit{Captioning with separate localization}} \\
Veason-R1-3B & $9.15_{\pm 0.37}$ & $11.58_{\pm 0.43}$ & $14.61_{\pm 0.05}$ & $7.74_{\pm 0.05}$ & $14.05_{\pm 0.04}$ & $10.97_{\pm 0.07}$ \\
InternVL3-8B + SAM2 & $9.72_{\pm 0.08}$ & $8.95_{\pm 0.43}$ & $18.55_{\pm 0.48}$ & $11.62_{\pm 0.15}$ & $11.01_{\pm 0.07}$ & $9.05_{\pm 0.18}$ \\
Qwen3-VL-8B + SAM2 & $10.65_{\pm 0.28}$ & $10.18_{\pm 0.50}$ & $31.99_{\pm 0.22}$ & $16.84_{\pm 0.22}$ & $22.40_{\pm 0.30}$ & $17.39_{\pm 0.12}$ \\
InternVL3.5-8B + SAM2 & $5.59_{\pm 0.28}$ & $10.26_{\pm 0.98}$ & $16.45_{\pm 0.22}$ & $13.14_{\pm 0.15}$ & $10.85_{\pm 0.11}$ & $8.88_{\pm 0.04}$ \\
Qwen3.5-9B + SAM2 & $11.41_{\pm 0.04}$ & $8.60_{\pm 0.25}$ & $42.65_{\pm 0.26}$ & $27.17_{\pm 0.01}$ & $30.32_{\pm 0.02}$ & $27.24_{\pm 0.14}$ \\
\bottomrule
\end{tabular}
\end{table*}

\begin{table*}[t]
\centering
\caption{Motion GCC@0.1 across source datasets.}
\label{tab:supp_dataset_motion}
\footnotesize
\setlength{\tabcolsep}{3pt}
\renewcommand{\arraystretch}{1.1}
\begin{tabular}{lcccccc}
\toprule
Model & Ped1 & Ped2 & Avenue & ShanghaiTech & UCF-Crime & MSAD \\
\midrule
\multicolumn{7}{l}{\textit{Joint caption and mask generation}} \\
Sa2VA-8B & $28.19_{\pm 0.21}$ & $25.44_{\pm 0.25}$ & $0.00_{\pm 0.00}$ & $0.00_{\pm 0.00}$ & $7.83_{\pm 0.05}$ & $0.16_{\pm 0.00}$ \\
VideoGLaMM-3.8B & $1.98_{\pm 0.08}$ & $6.49_{\pm 0.25}$ & $1.06_{\pm 0.00}$ & $4.65_{\pm 0.18}$ & $3.19_{\pm 0.02}$ & $3.67_{\pm 0.04}$ \\
UniPixel-7B & $6.44_{\pm 0.00}$ & $4.91_{\pm 0.25}$ & $16.52_{\pm 0.27}$ & $8.01_{\pm 0.07}$ & $2.27_{\pm 0.02}$ & $4.65_{\pm 0.05}$ \\
\midrule
\multicolumn{7}{l}{\textit{Captioning with separate localization}} \\
Veason-R1-3B & $17.37_{\pm 0.14}$ & $22.63_{\pm 0.00}$ & $21.17_{\pm 0.17}$ & $14.87_{\pm 0.03}$ & $10.78_{\pm 0.06}$ & $7.48_{\pm 0.09}$ \\
InternVL3-8B + SAM2 & $11.41_{\pm 0.08}$ & $12.98_{\pm 0.25}$ & $22.55_{\pm 0.35}$ & $17.44_{\pm 0.20}$ & $11.52_{\pm 0.06}$ & $11.05_{\pm 0.13}$ \\
Qwen3-VL-8B + SAM2 & $27.74_{\pm 0.28}$ & $44.12_{\pm 0.50}$ & $30.67_{\pm 0.22}$ & $28.54_{\pm 0.07}$ & $21.98_{\pm 0.07}$ & $18.00_{\pm 0.08}$ \\
InternVL3.5-8B + SAM2 & $8.42_{\pm 0.16}$ & $16.67_{\pm 0.25}$ & $16.13_{\pm 0.25}$ & $16.55_{\pm 0.14}$ & $9.85_{\pm 0.11}$ & $10.62_{\pm 0.13}$ \\
Qwen3.5-9B + SAM2 & $10.71_{\pm 0.08}$ & $16.84_{\pm 0.00}$ & $37.87_{\pm 0.43}$ & $34.26_{\pm 0.11}$ & $25.56_{\pm 0.09}$ & $24.08_{\pm 0.18}$ \\
\bottomrule
\end{tabular}
\end{table*}

\FloatBarrier

\subsection{Caption-Only Comparisons}
\label{sec:supp_caption_only}

Table~\ref{tab:supp_caption_only} reports caption similarity for the four VLM backbones without spatial evaluation.
For anomaly, these are the same event descriptions evaluated in the VLM+SAM2 rows of Table~\ref{tab:main_results}, so their Cap scores are identical.
Appearance and motion use separate generations that request descriptions without localization outputs.
Their Cap scores remain close to those of the corresponding pipelines in Table~\ref{tab:instance_results}, with differences of at most 1.33 points across the four backbones.
These comparisons concern description quality, while GCC evaluates whether the generated descriptions are supported by their associated tracks.

\begin{table}[t]
\centering
\caption{Caption-only Cap scores.}
\label{tab:supp_caption_only}
\footnotesize
\setlength{\tabcolsep}{3pt}
\renewcommand{\arraystretch}{1.1}
\begin{tabular}{lccc}
\toprule
Model & Anomaly & Appearance & Motion \\
\midrule
InternVL3-8B & $40.87_{\pm 0.12}$ & $24.56_{\pm 0.06}$ & $28.06_{\pm 0.02}$ \\
Qwen3-VL-8B & $43.98_{\pm 0.16}$ & $23.58_{\pm 0.04}$ & $29.70_{\pm 0.09}$ \\
InternVL3.5-8B & $37.26_{\pm 0.12}$ & $25.70_{\pm 0.08}$ & $28.33_{\pm 0.08}$ \\
Qwen3.5-9B & $45.54_{\pm 0.36}$ & $37.69_{\pm 0.05}$ & $34.95_{\pm 0.03}$ \\
\bottomrule
\end{tabular}
\end{table}

\FloatBarrier

\section{Robustness Analysis}
\label{sec:supp_robustness}

\FloatBarrier

\subsection{Grounding Thresholds}
\label{sec:supp_thresholds}

Table~\ref{tab:supp_thresholds} reports GCC at IoU thresholds of 0.1, 0.3, and 0.5, together with GCC@avg over the nine thresholds from 0.1 to 0.9.
We compute GCC@avg separately within each judge round, then report its mean and population standard deviation across the three rounds.
The two Qwen pipelines retain the top two positions under GCC@0.1 and GCC@avg in all three tasks.
Qwen3-VL leads on anomaly captions and Qwen3.5 leads on appearance and motion.
Veason-R1 ranks third under GCC@avg in all three tasks.
For anomaly captions, its GCC@0.1 is slightly lower than InternVL3.5+SAM2, while its GCC@avg is higher.

\begin{table*}[t]
\centering
\caption{Grounded caption scores across localization thresholds.}
\label{tab:supp_thresholds}
\footnotesize
\setlength{\tabcolsep}{0.5pt}
\renewcommand{\arraystretch}{1.1}
\begin{tabular}{lcccccccccccc}
\toprule
Model & \multicolumn{4}{c}{\textbf{Anomaly}} & \multicolumn{4}{c}{\textbf{Appearance}} & \multicolumn{4}{c}{\textbf{Motion}} \\
\cmidrule(lr){2-5} \cmidrule(lr){6-9} \cmidrule(lr){10-13}
 & GCC@0.1 & GCC@0.3 & GCC@0.5 & GCC@avg & GCC@0.1 & GCC@0.3 & GCC@0.5 & GCC@avg & GCC@0.1 & GCC@0.3 & GCC@0.5 & GCC@avg \\
\midrule
\multicolumn{13}{l}{\textit{Joint caption and mask generation}} \\
Sa2VA & \begin{tabular}[c]{@{}c@{}}$0.19$\\[-3pt]$\scriptstyle\pm 0.01$\end{tabular} & \begin{tabular}[c]{@{}c@{}}$0.19$\\[-3pt]$\scriptstyle\pm 0.01$\end{tabular} & \begin{tabular}[c]{@{}c@{}}$0.04$\\[-3pt]$\scriptstyle\pm 0.01$\end{tabular} & \begin{tabular}[c]{@{}c@{}}$0.10$\\[-3pt]$\scriptstyle\pm 0.01$\end{tabular} & \begin{tabular}[c]{@{}c@{}}$3.52$\\[-3pt]$\scriptstyle\pm 0.03$\end{tabular} & \begin{tabular}[c]{@{}c@{}}$3.06$\\[-3pt]$\scriptstyle\pm 0.03$\end{tabular} & \begin{tabular}[c]{@{}c@{}}$2.62$\\[-3pt]$\scriptstyle\pm 0.02$\end{tabular} & \begin{tabular}[c]{@{}c@{}}$2.43$\\[-3pt]$\scriptstyle\pm 0.02$\end{tabular} & \begin{tabular}[c]{@{}c@{}}$5.74$\\[-3pt]$\scriptstyle\pm 0.02$\end{tabular} & \begin{tabular}[c]{@{}c@{}}$4.93$\\[-3pt]$\scriptstyle\pm 0.01$\end{tabular} & \begin{tabular}[c]{@{}c@{}}$3.95$\\[-3pt]$\scriptstyle\pm 0.02$\end{tabular} & \begin{tabular}[c]{@{}c@{}}$3.69$\\[-3pt]$\scriptstyle\pm 0.01$\end{tabular} \\
VideoGLaMM-3.8B & \begin{tabular}[c]{@{}c@{}}$0.96$\\[-3pt]$\scriptstyle\pm 0.04$\end{tabular} & \begin{tabular}[c]{@{}c@{}}$0.70$\\[-3pt]$\scriptstyle\pm 0.03$\end{tabular} & \begin{tabular}[c]{@{}c@{}}$0.70$\\[-3pt]$\scriptstyle\pm 0.03$\end{tabular} & \begin{tabular}[c]{@{}c@{}}$0.68$\\[-3pt]$\scriptstyle\pm 0.03$\end{tabular} & \begin{tabular}[c]{@{}c@{}}$2.60$\\[-3pt]$\scriptstyle\pm 0.01$\end{tabular} & \begin{tabular}[c]{@{}c@{}}$2.39$\\[-3pt]$\scriptstyle\pm 0.01$\end{tabular} & \begin{tabular}[c]{@{}c@{}}$1.82$\\[-3pt]$\scriptstyle\pm 0.02$\end{tabular} & \begin{tabular}[c]{@{}c@{}}$1.87$\\[-3pt]$\scriptstyle\pm 0.01$\end{tabular} & \begin{tabular}[c]{@{}c@{}}$3.49$\\[-3pt]$\scriptstyle\pm 0.04$\end{tabular} & \begin{tabular}[c]{@{}c@{}}$2.94$\\[-3pt]$\scriptstyle\pm 0.04$\end{tabular} & \begin{tabular}[c]{@{}c@{}}$2.42$\\[-3pt]$\scriptstyle\pm 0.03$\end{tabular} & \begin{tabular}[c]{@{}c@{}}$2.47$\\[-3pt]$\scriptstyle\pm 0.03$\end{tabular} \\
UniPixel & \begin{tabular}[c]{@{}c@{}}$4.06$\\[-3pt]$\scriptstyle\pm 0.05$\end{tabular} & \begin{tabular}[c]{@{}c@{}}$3.45$\\[-3pt]$\scriptstyle\pm 0.05$\end{tabular} & \begin{tabular}[c]{@{}c@{}}$2.97$\\[-3pt]$\scriptstyle\pm 0.05$\end{tabular} & \begin{tabular}[c]{@{}c@{}}$2.79$\\[-3pt]$\scriptstyle\pm 0.04$\end{tabular} & \begin{tabular}[c]{@{}c@{}}$1.06$\\[-3pt]$\scriptstyle\pm 0.01$\end{tabular} & \begin{tabular}[c]{@{}c@{}}$0.87$\\[-3pt]$\scriptstyle\pm 0.01$\end{tabular} & \begin{tabular}[c]{@{}c@{}}$0.83$\\[-3pt]$\scriptstyle\pm 0.01$\end{tabular} & \begin{tabular}[c]{@{}c@{}}$0.79$\\[-3pt]$\scriptstyle\pm 0.01$\end{tabular} & \begin{tabular}[c]{@{}c@{}}$5.39$\\[-3pt]$\scriptstyle\pm 0.03$\end{tabular} & \begin{tabular}[c]{@{}c@{}}$4.29$\\[-3pt]$\scriptstyle\pm 0.02$\end{tabular} & \begin{tabular}[c]{@{}c@{}}$3.77$\\[-3pt]$\scriptstyle\pm 0.02$\end{tabular} & \begin{tabular}[c]{@{}c@{}}$3.64$\\[-3pt]$\scriptstyle\pm 0.02$\end{tabular} \\
\midrule
\multicolumn{13}{l}{\textit{Captioning with separate localization}} \\
Veason-R1-3B & \begin{tabular}[c]{@{}c@{}}$13.96$\\[-3pt]$\scriptstyle\pm 0.05$\end{tabular} & \begin{tabular}[c]{@{}c@{}}$12.22$\\[-3pt]$\scriptstyle\pm 0.04$\end{tabular} & \begin{tabular}[c]{@{}c@{}}$10.15$\\[-3pt]$\scriptstyle\pm 0.03$\end{tabular} & \begin{tabular}[c]{@{}c@{}}$9.50$\\[-3pt]$\scriptstyle\pm 0.04$\end{tabular} & \begin{tabular}[c]{@{}c@{}}$11.49$\\[-3pt]$\scriptstyle\pm 0.06$\end{tabular} & \begin{tabular}[c]{@{}c@{}}$10.86$\\[-3pt]$\scriptstyle\pm 0.08$\end{tabular} & \begin{tabular}[c]{@{}c@{}}$10.14$\\[-3pt]$\scriptstyle\pm 0.06$\end{tabular} & \begin{tabular}[c]{@{}c@{}}$9.41$\\[-3pt]$\scriptstyle\pm 0.05$\end{tabular} & \begin{tabular}[c]{@{}c@{}}$12.32$\\[-3pt]$\scriptstyle\pm 0.03$\end{tabular} & \begin{tabular}[c]{@{}c@{}}$10.88$\\[-3pt]$\scriptstyle\pm 0.02$\end{tabular} & \begin{tabular}[c]{@{}c@{}}$8.82$\\[-3pt]$\scriptstyle\pm 0.02$\end{tabular} & \begin{tabular}[c]{@{}c@{}}$8.82$\\[-3pt]$\scriptstyle\pm 0.02$\end{tabular} \\
InternVL3 + SAM2 & \begin{tabular}[c]{@{}c@{}}$19.54$\\[-3pt]$\scriptstyle\pm 0.12$\end{tabular} & \begin{tabular}[c]{@{}c@{}}$11.99$\\[-3pt]$\scriptstyle\pm 0.10$\end{tabular} & \begin{tabular}[c]{@{}c@{}}$6.58$\\[-3pt]$\scriptstyle\pm 0.05$\end{tabular} & \begin{tabular}[c]{@{}c@{}}$8.12$\\[-3pt]$\scriptstyle\pm 0.05$\end{tabular} & \begin{tabular}[c]{@{}c@{}}$10.98$\\[-3pt]$\scriptstyle\pm 0.07$\end{tabular} & \begin{tabular}[c]{@{}c@{}}$7.86$\\[-3pt]$\scriptstyle\pm 0.02$\end{tabular} & \begin{tabular}[c]{@{}c@{}}$5.15$\\[-3pt]$\scriptstyle\pm 0.02$\end{tabular} & \begin{tabular}[c]{@{}c@{}}$5.30$\\[-3pt]$\scriptstyle\pm 0.03$\end{tabular} & \begin{tabular}[c]{@{}c@{}}$13.42$\\[-3pt]$\scriptstyle\pm 0.09$\end{tabular} & \begin{tabular}[c]{@{}c@{}}$9.34$\\[-3pt]$\scriptstyle\pm 0.08$\end{tabular} & \begin{tabular}[c]{@{}c@{}}$5.74$\\[-3pt]$\scriptstyle\pm 0.03$\end{tabular} & \begin{tabular}[c]{@{}c@{}}$6.20$\\[-3pt]$\scriptstyle\pm 0.04$\end{tabular} \\
Qwen3-VL + SAM2 & \begin{tabular}[c]{@{}c@{}}$33.67$\\[-3pt]$\scriptstyle\pm 0.12$\end{tabular} & \begin{tabular}[c]{@{}c@{}}$29.52$\\[-3pt]$\scriptstyle\pm 0.11$\end{tabular} & \begin{tabular}[c]{@{}c@{}}$21.81$\\[-3pt]$\scriptstyle\pm 0.08$\end{tabular} & \begin{tabular}[c]{@{}c@{}}$20.85$\\[-3pt]$\scriptstyle\pm 0.08$\end{tabular} & \begin{tabular}[c]{@{}c@{}}$19.31$\\[-3pt]$\scriptstyle\pm 0.17$\end{tabular} & \begin{tabular}[c]{@{}c@{}}$18.15$\\[-3pt]$\scriptstyle\pm 0.17$\end{tabular} & \begin{tabular}[c]{@{}c@{}}$15.68$\\[-3pt]$\scriptstyle\pm 0.15$\end{tabular} & \begin{tabular}[c]{@{}c@{}}$14.60$\\[-3pt]$\scriptstyle\pm 0.13$\end{tabular} & \begin{tabular}[c]{@{}c@{}}$23.98$\\[-3pt]$\scriptstyle\pm 0.03$\end{tabular} & \begin{tabular}[c]{@{}c@{}}$21.67$\\[-3pt]$\scriptstyle\pm 0.04$\end{tabular} & \begin{tabular}[c]{@{}c@{}}$18.77$\\[-3pt]$\scriptstyle\pm 0.02$\end{tabular} & \begin{tabular}[c]{@{}c@{}}$17.20$\\[-3pt]$\scriptstyle\pm 0.03$\end{tabular} \\
InternVL3.5 + SAM2 & \begin{tabular}[c]{@{}c@{}}$14.31$\\[-3pt]$\scriptstyle\pm 0.04$\end{tabular} & \begin{tabular}[c]{@{}c@{}}$5.00$\\[-3pt]$\scriptstyle\pm 0.02$\end{tabular} & \begin{tabular}[c]{@{}c@{}}$1.93$\\[-3pt]$\scriptstyle\pm 0.01$\end{tabular} & \begin{tabular}[c]{@{}c@{}}$3.88$\\[-3pt]$\scriptstyle\pm 0.01$\end{tabular} & \begin{tabular}[c]{@{}c@{}}$10.80$\\[-3pt]$\scriptstyle\pm 0.09$\end{tabular} & \begin{tabular}[c]{@{}c@{}}$4.79$\\[-3pt]$\scriptstyle\pm 0.06$\end{tabular} & \begin{tabular}[c]{@{}c@{}}$2.03$\\[-3pt]$\scriptstyle\pm 0.04$\end{tabular} & \begin{tabular}[c]{@{}c@{}}$3.41$\\[-3pt]$\scriptstyle\pm 0.04$\end{tabular} & \begin{tabular}[c]{@{}c@{}}$11.97$\\[-3pt]$\scriptstyle\pm 0.04$\end{tabular} & \begin{tabular}[c]{@{}c@{}}$5.69$\\[-3pt]$\scriptstyle\pm 0.01$\end{tabular} & \begin{tabular}[c]{@{}c@{}}$2.37$\\[-3pt]$\scriptstyle\pm 0.02$\end{tabular} & \begin{tabular}[c]{@{}c@{}}$3.87$\\[-3pt]$\scriptstyle\pm 0.01$\end{tabular} \\
Qwen3.5 + SAM2 & \begin{tabular}[c]{@{}c@{}}$33.21$\\[-3pt]$\scriptstyle\pm 0.24$\end{tabular} & \begin{tabular}[c]{@{}c@{}}$26.38$\\[-3pt]$\scriptstyle\pm 0.20$\end{tabular} & \begin{tabular}[c]{@{}c@{}}$18.66$\\[-3pt]$\scriptstyle\pm 0.15$\end{tabular} & \begin{tabular}[c]{@{}c@{}}$18.55$\\[-3pt]$\scriptstyle\pm 0.16$\end{tabular} & \begin{tabular}[c]{@{}c@{}}$27.62$\\[-3pt]$\scriptstyle\pm 0.02$\end{tabular} & \begin{tabular}[c]{@{}c@{}}$24.99$\\[-3pt]$\scriptstyle\pm 0.02$\end{tabular} & \begin{tabular}[c]{@{}c@{}}$20.23$\\[-3pt]$\scriptstyle\pm 0.00$\end{tabular} & \begin{tabular}[c]{@{}c@{}}$19.46$\\[-3pt]$\scriptstyle\pm 0.01$\end{tabular} & \begin{tabular}[c]{@{}c@{}}$26.51$\\[-3pt]$\scriptstyle\pm 0.03$\end{tabular} & \begin{tabular}[c]{@{}c@{}}$22.62$\\[-3pt]$\scriptstyle\pm 0.04$\end{tabular} & \begin{tabular}[c]{@{}c@{}}$18.09$\\[-3pt]$\scriptstyle\pm 0.04$\end{tabular} & \begin{tabular}[c]{@{}c@{}}$17.47$\\[-3pt]$\scriptstyle\pm 0.04$\end{tabular} \\
\bottomrule
\end{tabular}
\end{table*}

Table~\ref{tab:supp_rgcc} evaluates anomaly captions with the model's declared roles.
rGCC requires the group and member IoUs to pass for every annotated role.
rGCC-strict additionally requires the predicted set of roles to equal the annotated set.
For example, a prediction that adds an affected participant to an event with only an acting participant receives zero rGCC-strict.
Qwen3-VL+SAM2 obtains rGCC of 24.24 and rGCC-strict of 12.10 at a threshold of 0.1.
The difference shows that predicting whether an affected participant is present remains difficult even when the annotated participants are localized.

\begin{table*}[t]
\centering
\caption{Role-aware grounding of anomaly captions across localization thresholds.}
\label{tab:supp_rgcc}
\footnotesize
\setlength{\tabcolsep}{3pt}
\renewcommand{\arraystretch}{1.1}
\begin{tabular}{lcccccc}
\toprule
Model & rGCC@0.1 & rGCC@0.3 & rGCC@0.5 & rGCC@avg & rGCC-strict@0.1 & rGCC-strict@avg \\
\midrule
\multicolumn{7}{l}{\textit{Joint caption and mask generation}} \\
Sa2VA-8B & $0.00_{\pm 0.00}$ & $0.00_{\pm 0.00}$ & $0.00_{\pm 0.00}$ & $0.00_{\pm 0.00}$ & $0.00_{\pm 0.00}$ & $0.00_{\pm 0.00}$ \\
VideoGLaMM-3.8B & $0.00_{\pm 0.00}$ & $0.00_{\pm 0.00}$ & $0.00_{\pm 0.00}$ & $0.00_{\pm 0.00}$ & $0.00_{\pm 0.00}$ & $0.00_{\pm 0.00}$ \\
UniPixel-7B & $0.00_{\pm 0.00}$ & $0.00_{\pm 0.00}$ & $0.00_{\pm 0.00}$ & $0.00_{\pm 0.00}$ & $0.00_{\pm 0.00}$ & $0.00_{\pm 0.00}$ \\
\midrule
\multicolumn{7}{l}{\textit{Captioning with separate localization}} \\
Veason-R1-3B & $8.61_{\pm 0.02}$ & $7.58_{\pm 0.02}$ & $6.86_{\pm 0.01}$ & $6.36_{\pm 0.01}$ & $6.31_{\pm 0.02}$ & $4.85_{\pm 0.01}$ \\
InternVL3-8B + SAM2 & $12.15_{\pm 0.11}$ & $7.96_{\pm 0.07}$ & $5.12_{\pm 0.03}$ & $5.56_{\pm 0.04}$ & $4.77_{\pm 0.03}$ & $1.96_{\pm 0.01}$ \\
Qwen3-VL-8B + SAM2 & $24.24_{\pm 0.11}$ & $21.61_{\pm 0.11}$ & $17.48_{\pm 0.13}$ & $16.02_{\pm 0.10}$ & $12.10_{\pm 0.05}$ & $7.66_{\pm 0.04}$ \\
InternVL3.5-8B + SAM2 & $6.76_{\pm 0.05}$ & $2.69_{\pm 0.02}$ & $1.04_{\pm 0.00}$ & $2.01_{\pm 0.01}$ & $4.08_{\pm 0.04}$ & $1.15_{\pm 0.01}$ \\
Qwen3.5-9B + SAM2 & $24.68_{\pm 0.12}$ & $18.31_{\pm 0.12}$ & $13.37_{\pm 0.15}$ & $13.70_{\pm 0.12}$ & $11.71_{\pm 0.04}$ & $6.24_{\pm 0.03}$ \\
\bottomrule
\end{tabular}
\end{table*}

\FloatBarrier

\subsection{Alternative Caption Judges}
\label{sec:supp_caption_judges}

Table~\ref{tab:supp_judges} gives caption and joint scores from Mistral Small 3.2 at temperature zero, together with METEOR caption scores.
With the Mistral judge, Qwen3.5+SAM2 again has a higher anomaly Cap than Qwen3-VL+SAM2 but lower GCC.
Qwen3.5+SAM2 has the higher rGCC@0.1 with both judges.

\begin{table*}[t]
\centering
\caption{Mistral caption evaluation and METEOR scores.}
\label{tab:supp_judges}
\footnotesize
\setlength{\tabcolsep}{3pt}
\renewcommand{\arraystretch}{1.1}
\begin{tabular}{lcccccccccc}
\toprule
Model & \multicolumn{4}{c}{\textbf{Anomaly}} & \multicolumn{3}{c}{\textbf{Appearance}} & \multicolumn{3}{c}{\textbf{Motion}} \\
\cmidrule(lr){2-5} \cmidrule(lr){6-8} \cmidrule(lr){9-11}
 & Cap & GCC@0.1 & rGCC@0.1 & METEOR & Cap & GCC@0.1 & METEOR & Cap & GCC@0.1 & METEOR \\
\midrule
\multicolumn{11}{l}{\textit{Joint caption and mask generation}} \\
Sa2VA-8B & 4.93 & 0.22 & 0.00 & 1.19 & 5.47 & 4.61 & 1.16 & 9.11 & 7.89 & 1.86 \\
VideoGLaMM-3.8B & 7.38 & 2.40 & 0.00 & 2.18 & 12.74 & 5.40 & 2.50 & 16.29 & 8.66 & 2.52 \\
UniPixel-7B & 30.34 & 5.19 & 0.00 & 6.34 & 1.86 & 1.05 & 0.35 & 16.89 & 7.25 & 2.81 \\
\midrule
\multicolumn{11}{l}{\textit{Captioning with separate localization}} \\
Veason-R1-3B & 34.93 & 17.23 & 10.58 & 8.05 & 19.42 & 13.77 & 4.93 & 26.71 & 16.53 & 6.02 \\
InternVL3-8B + SAM2 & 48.57 & 22.45 & 13.62 & 8.97 & 30.28 & 14.59 & 6.44 & 37.25 & 18.57 & 5.08 \\
Qwen3-VL-8B + SAM2 & 49.42 & 35.51 & 24.64 & 9.48 & 29.16 & 23.36 & 7.40 & 36.41 & 29.31 & 8.87 \\
InternVL3.5-8B + SAM2 & 45.95 & 16.53 & 7.38 & 8.60 & 31.57 & 14.80 & 5.83 & 39.48 & 17.31 & 7.08 \\
Qwen3.5-9B + SAM2 & 50.46 & 34.78 & 24.98 & 11.33 & 41.72 & 32.04 & 10.69 & 41.61 & 33.05 & 10.10 \\
\bottomrule
\end{tabular}
\end{table*}

\FloatBarrier

\subsection{Temporal Aggregation}
\label{sec:supp_temporal_aggregation}

Tracks with low temporal occupancy leave more frames in which both masks can be empty.
We define occupancy $\rho$ as the percentage of annotated frames with a nonempty ground-truth mask.
Mean occupancy is 84.2\% for instance tracks and 92.4\% for role-group tracks.
Table~\ref{tab:supp_occupancy} reports the distributions separately because each role-group track combines the masks of all participants with that role.
Individual tracks more often occupy only a small part of the clip.

\begin{table}[t]
\centering
\caption{Temporal occupancy of ground-truth tracks. Percentages are computed within each track type.}
\label{tab:supp_occupancy}
\small
\setlength{\tabcolsep}{3pt}
\renewcommand{\arraystretch}{1.1}
\begin{tabular}{lrr}
\toprule
Occupancy $\rho$ (\%) & Instance (\%) & Role group (\%) \\
\midrule
$0 < \rho \leq 25$ & 5.2 & 2.1 \\
$25 < \rho \leq 50$ & 8.1 & 2.7 \\
$50 < \rho \leq 75$ & 11.1 & 7.5 \\
$75 < \rho < 100$ & 24.0 & 21.4 \\
$\rho = 100$ & 51.5 & 66.3 \\
\bottomrule
\end{tabular}
\end{table}

Table~\ref{tab:supp_volume_gcc} compares GCC under frame-averaged track IoU and spatiotemporal volume IoU:
\begin{equation}
  \mathrm{IoU}_{\mathrm{vol}}(P,G)
  = \frac{\sum_{t\in\mathcal{T}} |P_t\cap G_t|}
           {\sum_{t\in\mathcal{T}} |P_t\cup G_t|},
  \label{eq:volume_iou}
\end{equation}
where $P_t$ and $G_t$ are the predicted and ground-truth masks over the same annotated frames $\mathcal{T}$.
Jointly empty frames contribute nothing to either sum, while predictions on frames without the instance increase the union.
Volume IoU also weights frames by their union area, whereas frame-averaged IoU weights each frame equally.

We recompute GCC with volume IoU for the eight models with mask outputs in Tables~\ref{tab:main_results} and~\ref{tab:instance_results}.
Predictions, pairwise caption similarity scores, and evaluation frames remain fixed.
Hungarian matching is recomputed under each IoU definition.
For appearance and motion, the caption score follows the prediction selected by the new mask assignment.
Both variants use the same thresholds and pooled-micro aggregation.
We report means and population standard deviations across the same three judge rounds.

For rGCC, we hold the declared roles and local false-positive assignments fixed and compare frame averaging with volume aggregation (Table~\ref{tab:supp_volume_rgcc}).
The volume variant sums each member's intersections and denominators in Eq.~\ref{eq:member_iou} over its visible frames before taking their ratio.
Role-group overlap uses Eq.~\ref{eq:volume_iou} over the full clip.
Both variants require every group and member to pass the same threshold, and neither credits jointly empty frames.

Table~\ref{tab:supp_volume_gcc} shows that the caption-grounding discrepancy persists under volume IoU.
Qwen3.5+SAM2 retains higher anomaly Cap than Qwen3-VL+SAM2 but lower GCC@0.1.
InternVL3+SAM2 and Qwen3-VL+SAM2 also retain nearly identical appearance Cap but substantially different joint scores.
Affected-instance GCC@0.1 remains lower than acting-instance GCC@0.1 in both appearance and motion for all eight models.

Model rankings agree closely at GCC@0.1, with Kendall's $\tau_b$ of 0.929 for anomaly and 1.000 for both appearance and motion.
The anomaly change is a swap between Veason-R1 and InternVL3.5+SAM2.
Table~\ref{tab:supp_volume_rgcc} shows that volume aggregation reverses the two Qwen models' order under rGCC@0.1, while rGCC@avg rankings remain unchanged.
For GCC@avg, the anomaly and appearance rankings are unchanged, while motion has $\tau_b=0.786$.
Qwen3.5+SAM2 leads Qwen3-VL+SAM2 under frame-averaged motion GCC@avg but falls below it under volume IoU.
Temporal aggregation therefore affects some model orderings, including the leading motion model under GCC@avg, while the caption-grounding discrepancy and lower affected-instance GCC persist under both definitions.

\begin{table*}[t]
\centering
\caption{GCC under frame-averaged and volume IoU.}
\label{tab:supp_volume_gcc}
\footnotesize
\setlength{\tabcolsep}{2pt}
\renewcommand{\arraystretch}{1.1}
\begin{tabular}{lcccccccccccc}
\toprule
Model & \multicolumn{4}{c}{\textbf{Anomaly}} & \multicolumn{4}{c}{\textbf{Appearance}} & \multicolumn{4}{c}{\textbf{Motion}} \\
\cmidrule(lr){2-5} \cmidrule(lr){6-9} \cmidrule(lr){10-13}
 & \multicolumn{2}{c}{GCC@0.1} & \multicolumn{2}{c}{GCC@avg} & \multicolumn{2}{c}{GCC@0.1} & \multicolumn{2}{c}{GCC@avg} & \multicolumn{2}{c}{GCC@0.1} & \multicolumn{2}{c}{GCC@avg} \\
\cmidrule(lr){2-3} \cmidrule(lr){4-5} \cmidrule(lr){6-7} \cmidrule(lr){8-9} \cmidrule(lr){10-11} \cmidrule(lr){12-13}
 & Frame & Volume & Frame & Volume & Frame & Volume & Frame & Volume & Frame & Volume & Frame & Volume \\
\midrule
\multicolumn{13}{l}{\textit{Joint caption and mask generation}} \\
Sa2VA & \begin{tabular}[c]{@{}c@{}}$0.19$\\[-3pt]$\scriptstyle\pm 0.01$\end{tabular} & \begin{tabular}[c]{@{}c@{}}$0.19$\\[-3pt]$\scriptstyle\pm 0.01$\end{tabular} & \begin{tabular}[c]{@{}c@{}}$0.10$\\[-3pt]$\scriptstyle\pm 0.01$\end{tabular} & \begin{tabular}[c]{@{}c@{}}$0.09$\\[-3pt]$\scriptstyle\pm 0.01$\end{tabular} & \begin{tabular}[c]{@{}c@{}}$3.52$\\[-3pt]$\scriptstyle\pm 0.03$\end{tabular} & \begin{tabular}[c]{@{}c@{}}$3.49$\\[-3pt]$\scriptstyle\pm 0.03$\end{tabular} & \begin{tabular}[c]{@{}c@{}}$2.43$\\[-3pt]$\scriptstyle\pm 0.02$\end{tabular} & \begin{tabular}[c]{@{}c@{}}$2.40$\\[-3pt]$\scriptstyle\pm 0.03$\end{tabular} & \begin{tabular}[c]{@{}c@{}}$5.74$\\[-3pt]$\scriptstyle\pm 0.02$\end{tabular} & \begin{tabular}[c]{@{}c@{}}$5.57$\\[-3pt]$\scriptstyle\pm 0.02$\end{tabular} & \begin{tabular}[c]{@{}c@{}}$3.69$\\[-3pt]$\scriptstyle\pm 0.01$\end{tabular} & \begin{tabular}[c]{@{}c@{}}$3.59$\\[-3pt]$\scriptstyle\pm 0.01$\end{tabular} \\
VideoGLaMM-3.8B & \begin{tabular}[c]{@{}c@{}}$0.96$\\[-3pt]$\scriptstyle\pm 0.04$\end{tabular} & \begin{tabular}[c]{@{}c@{}}$0.96$\\[-3pt]$\scriptstyle\pm 0.04$\end{tabular} & \begin{tabular}[c]{@{}c@{}}$0.68$\\[-3pt]$\scriptstyle\pm 0.03$\end{tabular} & \begin{tabular}[c]{@{}c@{}}$0.69$\\[-3pt]$\scriptstyle\pm 0.03$\end{tabular} & \begin{tabular}[c]{@{}c@{}}$2.60$\\[-3pt]$\scriptstyle\pm 0.01$\end{tabular} & \begin{tabular}[c]{@{}c@{}}$2.53$\\[-3pt]$\scriptstyle\pm 0.01$\end{tabular} & \begin{tabular}[c]{@{}c@{}}$1.87$\\[-3pt]$\scriptstyle\pm 0.01$\end{tabular} & \begin{tabular}[c]{@{}c@{}}$1.80$\\[-3pt]$\scriptstyle\pm 0.01$\end{tabular} & \begin{tabular}[c]{@{}c@{}}$3.49$\\[-3pt]$\scriptstyle\pm 0.04$\end{tabular} & \begin{tabular}[c]{@{}c@{}}$3.09$\\[-3pt]$\scriptstyle\pm 0.02$\end{tabular} & \begin{tabular}[c]{@{}c@{}}$2.47$\\[-3pt]$\scriptstyle\pm 0.03$\end{tabular} & \begin{tabular}[c]{@{}c@{}}$2.29$\\[-3pt]$\scriptstyle\pm 0.03$\end{tabular} \\
UniPixel-7B & \begin{tabular}[c]{@{}c@{}}$4.06$\\[-3pt]$\scriptstyle\pm 0.05$\end{tabular} & \begin{tabular}[c]{@{}c@{}}$4.06$\\[-3pt]$\scriptstyle\pm 0.05$\end{tabular} & \begin{tabular}[c]{@{}c@{}}$2.79$\\[-3pt]$\scriptstyle\pm 0.04$\end{tabular} & \begin{tabular}[c]{@{}c@{}}$2.82$\\[-3pt]$\scriptstyle\pm 0.04$\end{tabular} & \begin{tabular}[c]{@{}c@{}}$1.06$\\[-3pt]$\scriptstyle\pm 0.01$\end{tabular} & \begin{tabular}[c]{@{}c@{}}$1.03$\\[-3pt]$\scriptstyle\pm 0.01$\end{tabular} & \begin{tabular}[c]{@{}c@{}}$0.79$\\[-3pt]$\scriptstyle\pm 0.01$\end{tabular} & \begin{tabular}[c]{@{}c@{}}$0.75$\\[-3pt]$\scriptstyle\pm 0.01$\end{tabular} & \begin{tabular}[c]{@{}c@{}}$5.39$\\[-3pt]$\scriptstyle\pm 0.03$\end{tabular} & \begin{tabular}[c]{@{}c@{}}$5.19$\\[-3pt]$\scriptstyle\pm 0.03$\end{tabular} & \begin{tabular}[c]{@{}c@{}}$3.64$\\[-3pt]$\scriptstyle\pm 0.02$\end{tabular} & \begin{tabular}[c]{@{}c@{}}$3.55$\\[-3pt]$\scriptstyle\pm 0.02$\end{tabular} \\
\midrule
\multicolumn{13}{l}{\textit{Captioning with separate localization}} \\
Veason-R1-3B & \begin{tabular}[c]{@{}c@{}}$13.96$\\[-3pt]$\scriptstyle\pm 0.05$\end{tabular} & \begin{tabular}[c]{@{}c@{}}$13.83$\\[-3pt]$\scriptstyle\pm 0.04$\end{tabular} & \begin{tabular}[c]{@{}c@{}}$9.50$\\[-3pt]$\scriptstyle\pm 0.04$\end{tabular} & \begin{tabular}[c]{@{}c@{}}$9.41$\\[-3pt]$\scriptstyle\pm 0.04$\end{tabular} & \begin{tabular}[c]{@{}c@{}}$11.49$\\[-3pt]$\scriptstyle\pm 0.06$\end{tabular} & \begin{tabular}[c]{@{}c@{}}$11.31$\\[-3pt]$\scriptstyle\pm 0.08$\end{tabular} & \begin{tabular}[c]{@{}c@{}}$9.41$\\[-3pt]$\scriptstyle\pm 0.05$\end{tabular} & \begin{tabular}[c]{@{}c@{}}$9.47$\\[-3pt]$\scriptstyle\pm 0.05$\end{tabular} & \begin{tabular}[c]{@{}c@{}}$12.32$\\[-3pt]$\scriptstyle\pm 0.03$\end{tabular} & \begin{tabular}[c]{@{}c@{}}$12.18$\\[-3pt]$\scriptstyle\pm 0.03$\end{tabular} & \begin{tabular}[c]{@{}c@{}}$8.82$\\[-3pt]$\scriptstyle\pm 0.02$\end{tabular} & \begin{tabular}[c]{@{}c@{}}$8.54$\\[-3pt]$\scriptstyle\pm 0.02$\end{tabular} \\
InternVL3 + SAM2 & \begin{tabular}[c]{@{}c@{}}$19.54$\\[-3pt]$\scriptstyle\pm 0.12$\end{tabular} & \begin{tabular}[c]{@{}c@{}}$16.90$\\[-3pt]$\scriptstyle\pm 0.08$\end{tabular} & \begin{tabular}[c]{@{}c@{}}$8.12$\\[-3pt]$\scriptstyle\pm 0.05$\end{tabular} & \begin{tabular}[c]{@{}c@{}}$7.29$\\[-3pt]$\scriptstyle\pm 0.04$\end{tabular} & \begin{tabular}[c]{@{}c@{}}$10.98$\\[-3pt]$\scriptstyle\pm 0.07$\end{tabular} & \begin{tabular}[c]{@{}c@{}}$10.67$\\[-3pt]$\scriptstyle\pm 0.04$\end{tabular} & \begin{tabular}[c]{@{}c@{}}$5.30$\\[-3pt]$\scriptstyle\pm 0.03$\end{tabular} & \begin{tabular}[c]{@{}c@{}}$4.50$\\[-3pt]$\scriptstyle\pm 0.02$\end{tabular} & \begin{tabular}[c]{@{}c@{}}$13.42$\\[-3pt]$\scriptstyle\pm 0.09$\end{tabular} & \begin{tabular}[c]{@{}c@{}}$12.55$\\[-3pt]$\scriptstyle\pm 0.07$\end{tabular} & \begin{tabular}[c]{@{}c@{}}$6.20$\\[-3pt]$\scriptstyle\pm 0.04$\end{tabular} & \begin{tabular}[c]{@{}c@{}}$5.11$\\[-3pt]$\scriptstyle\pm 0.03$\end{tabular} \\
Qwen3-VL + SAM2 & \begin{tabular}[c]{@{}c@{}}$33.67$\\[-3pt]$\scriptstyle\pm 0.12$\end{tabular} & \begin{tabular}[c]{@{}c@{}}$33.03$\\[-3pt]$\scriptstyle\pm 0.09$\end{tabular} & \begin{tabular}[c]{@{}c@{}}$20.85$\\[-3pt]$\scriptstyle\pm 0.08$\end{tabular} & \begin{tabular}[c]{@{}c@{}}$20.84$\\[-3pt]$\scriptstyle\pm 0.09$\end{tabular} & \begin{tabular}[c]{@{}c@{}}$19.31$\\[-3pt]$\scriptstyle\pm 0.17$\end{tabular} & \begin{tabular}[c]{@{}c@{}}$19.17$\\[-3pt]$\scriptstyle\pm 0.18$\end{tabular} & \begin{tabular}[c]{@{}c@{}}$14.60$\\[-3pt]$\scriptstyle\pm 0.13$\end{tabular} & \begin{tabular}[c]{@{}c@{}}$14.19$\\[-3pt]$\scriptstyle\pm 0.13$\end{tabular} & \begin{tabular}[c]{@{}c@{}}$23.98$\\[-3pt]$\scriptstyle\pm 0.03$\end{tabular} & \begin{tabular}[c]{@{}c@{}}$23.64$\\[-3pt]$\scriptstyle\pm 0.05$\end{tabular} & \begin{tabular}[c]{@{}c@{}}$17.20$\\[-3pt]$\scriptstyle\pm 0.03$\end{tabular} & \begin{tabular}[c]{@{}c@{}}$16.62$\\[-3pt]$\scriptstyle\pm 0.03$\end{tabular} \\
InternVL3.5 + SAM2 & \begin{tabular}[c]{@{}c@{}}$14.31$\\[-3pt]$\scriptstyle\pm 0.04$\end{tabular} & \begin{tabular}[c]{@{}c@{}}$12.18$\\[-3pt]$\scriptstyle\pm 0.06$\end{tabular} & \begin{tabular}[c]{@{}c@{}}$3.88$\\[-3pt]$\scriptstyle\pm 0.01$\end{tabular} & \begin{tabular}[c]{@{}c@{}}$3.33$\\[-3pt]$\scriptstyle\pm 0.01$\end{tabular} & \begin{tabular}[c]{@{}c@{}}$10.80$\\[-3pt]$\scriptstyle\pm 0.09$\end{tabular} & \begin{tabular}[c]{@{}c@{}}$9.44$\\[-3pt]$\scriptstyle\pm 0.07$\end{tabular} & \begin{tabular}[c]{@{}c@{}}$3.41$\\[-3pt]$\scriptstyle\pm 0.04$\end{tabular} & \begin{tabular}[c]{@{}c@{}}$2.65$\\[-3pt]$\scriptstyle\pm 0.03$\end{tabular} & \begin{tabular}[c]{@{}c@{}}$11.97$\\[-3pt]$\scriptstyle\pm 0.04$\end{tabular} & \begin{tabular}[c]{@{}c@{}}$10.05$\\[-3pt]$\scriptstyle\pm 0.05$\end{tabular} & \begin{tabular}[c]{@{}c@{}}$3.87$\\[-3pt]$\scriptstyle\pm 0.01$\end{tabular} & \begin{tabular}[c]{@{}c@{}}$2.86$\\[-3pt]$\scriptstyle\pm 0.01$\end{tabular} \\
Qwen3.5 + SAM2 & \begin{tabular}[c]{@{}c@{}}$33.21$\\[-3pt]$\scriptstyle\pm 0.24$\end{tabular} & \begin{tabular}[c]{@{}c@{}}$32.15$\\[-3pt]$\scriptstyle\pm 0.21$\end{tabular} & \begin{tabular}[c]{@{}c@{}}$18.55$\\[-3pt]$\scriptstyle\pm 0.16$\end{tabular} & \begin{tabular}[c]{@{}c@{}}$18.02$\\[-3pt]$\scriptstyle\pm 0.14$\end{tabular} & \begin{tabular}[c]{@{}c@{}}$27.62$\\[-3pt]$\scriptstyle\pm 0.02$\end{tabular} & \begin{tabular}[c]{@{}c@{}}$26.29$\\[-3pt]$\scriptstyle\pm 0.02$\end{tabular} & \begin{tabular}[c]{@{}c@{}}$19.46$\\[-3pt]$\scriptstyle\pm 0.01$\end{tabular} & \begin{tabular}[c]{@{}c@{}}$18.17$\\[-3pt]$\scriptstyle\pm 0.02$\end{tabular} & \begin{tabular}[c]{@{}c@{}}$26.51$\\[-3pt]$\scriptstyle\pm 0.03$\end{tabular} & \begin{tabular}[c]{@{}c@{}}$24.79$\\[-3pt]$\scriptstyle\pm 0.05$\end{tabular} & \begin{tabular}[c]{@{}c@{}}$17.47$\\[-3pt]$\scriptstyle\pm 0.04$\end{tabular} & \begin{tabular}[c]{@{}c@{}}$16.12$\\[-3pt]$\scriptstyle\pm 0.05$\end{tabular} \\
\bottomrule
\end{tabular}
\end{table*}

\begin{table*}[t]
\centering
\caption{Role-aware GCC with frame and volume aggregation.}
\label{tab:supp_volume_rgcc}
\footnotesize
\setlength{\tabcolsep}{3pt}
\renewcommand{\arraystretch}{1.1}
\begin{tabular}{lcccc}
\toprule
Model & \multicolumn{2}{c}{rGCC@0.1} & \multicolumn{2}{c}{rGCC@avg} \\
\cmidrule(lr){2-3} \cmidrule(lr){4-5}
 & Frame & Volume & Frame & Volume \\
\midrule
\multicolumn{5}{l}{\textit{Joint caption and mask generation}} \\
Sa2VA & $0.00_{\pm 0.00}$ & $0.00_{\pm 0.00}$ & $0.00_{\pm 0.00}$ & $0.00_{\pm 0.00}$ \\
VideoGLaMM-3.8B & $0.00_{\pm 0.00}$ & $0.00_{\pm 0.00}$ & $0.00_{\pm 0.00}$ & $0.00_{\pm 0.00}$ \\
UniPixel-7B & $0.00_{\pm 0.00}$ & $0.00_{\pm 0.00}$ & $0.00_{\pm 0.00}$ & $0.00_{\pm 0.00}$ \\
\midrule
\multicolumn{5}{l}{\textit{Captioning with separate localization}} \\
Veason-R1-3B & $8.61_{\pm 0.02}$ & $8.55_{\pm 0.02}$ & $6.36_{\pm 0.01}$ & $6.43_{\pm 0.01}$ \\
InternVL3 + SAM2 & $12.15_{\pm 0.11}$ & $11.46_{\pm 0.09}$ & $5.56_{\pm 0.04}$ & $5.31_{\pm 0.04}$ \\
Qwen3-VL + SAM2 & $24.24_{\pm 0.11}$ & $24.31_{\pm 0.09}$ & $16.02_{\pm 0.10}$ & $16.46_{\pm 0.09}$ \\
InternVL3.5 + SAM2 & $6.76_{\pm 0.05}$ & $7.04_{\pm 0.05}$ & $2.01_{\pm 0.01}$ & $1.99_{\pm 0.01}$ \\
Qwen3.5 + SAM2 & $24.68_{\pm 0.12}$ & $24.07_{\pm 0.12}$ & $13.70_{\pm 0.12}$ & $13.62_{\pm 0.11}$ \\
\bottomrule
\end{tabular}
\end{table*}

\FloatBarrier

\section{Participant Analyses}
\label{sec:supp_participant_analyses}

\FloatBarrier

\subsection{Group and Individual Participant Grounding}
\label{sec:supp_member_grounding}

Table~\ref{tab:supp_member_grounding} isolates the effect of checking individual participants when grounding anomaly captions.
Both columns report success rates over the same subset of clips containing multiple instances of at least one role, which accounts for 34.0\% of the benchmark.
The role-group condition requires every annotated role group's IoU to reach 0.1.
The combined condition additionally requires each participant's local IoU to reach the same threshold.
Predictions, declared roles, and frame-averaging rules in Eq.~\ref{eq:member_temporal} are held fixed.

Every model with nonzero role-group success has a lower pass rate when each participant is checked.
Among clips where Qwen3-VL+SAM2 passes the role-group condition, 73.2\% fail the individual-participant check.
This conditional rate uses group-passing clips as its denominator, while both table columns use the full subset.
These results show that sufficient role-group overlap can coexist with inadequate localization of individual participants.

\begin{table*}[t]
\centering
\caption{Effect of participant checks when grounding anomaly captions.}
\label{tab:supp_member_grounding}
\small
\setlength{\tabcolsep}{3pt}
\renewcommand{\arraystretch}{1.1}
\begin{tabular}{lrr}
\toprule
Model & \shortstack{Role-group\\overlap (\%)} & \shortstack{Role-group and\\individual-participant\\overlap (\%)} \\
\midrule
\multicolumn{3}{l}{\textit{Joint caption and mask generation}} \\
Sa2VA-8B & 0.0 & 0.0 \\
VideoGLaMM-3.8B & 0.0 & 0.0 \\
UniPixel-7B & 0.0 & 0.0 \\
\midrule
\multicolumn{3}{l}{\textit{Captioning with separate localization}} \\
Veason-R1-3B & 38.6 & 2.9 \\
InternVL3-8B + SAM2 & 36.4 & 9.3 \\
Qwen3-VL-8B + SAM2 & 58.6 & 15.7 \\
InternVL3.5-8B + SAM2 & 25.0 & 4.3 \\
Qwen3.5-9B + SAM2 & 50.7 & 14.3 \\
\bottomrule
\end{tabular}
\end{table*}

\FloatBarrier

\subsection{Participant Composition}
\label{sec:supp_participant_composition}

\begin{figure*}[t]
  \centering
  \includegraphics[width=\textwidth]{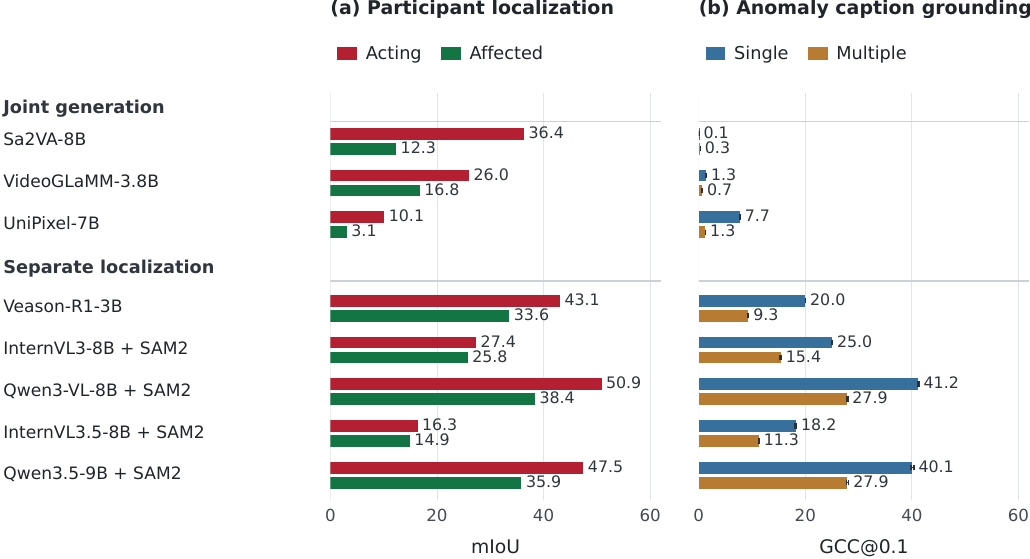}
  \caption{Acting and affected mIoU over all annotated tracks of each role (left), and anomaly GCC@0.1 by participant count (right). Error bars show judge standard deviations.}
  \label{fig:participant_grounding}
\end{figure*}

Figure~\ref{fig:participant_grounding} (left) reports localization over all annotated role-group tracks of each role, as in Table~\ref{tab:main_results}.
Affected role groups have lower mIoU in all eight systems.
Figure~\ref{fig:participant_grounding} (right) compares anomaly GCC between clips with one annotated instance and those with multiple instances.

Table~\ref{tab:supp_participant_composition} compares grounding of anomaly captions on single-instance and multiple-instance clips for Qwen3.5-9B+SAM2.
Clip percentages are computed within each source dataset and over the full benchmark in the pooled row.
The multiple-instance GCC is lower in all six source datasets.
The groups also differ in scene and event composition, so this comparison describes an association with participant count.

\begin{table*}[t]
\centering
\caption{Grounding of anomaly captions by participant count for Qwen3.5-9B+SAM2.}
\label{tab:supp_participant_composition}
\footnotesize
\setlength{\tabcolsep}{3pt}
\renewcommand{\arraystretch}{1.1}
\begin{tabular}{lrcrc}
\toprule
\multirow{2}{*}{Dataset} & \multicolumn{2}{c}{Single instance} & \multicolumn{2}{c}{Multiple instances} \\
\cmidrule(lr){2-3} \cmidrule(lr){4-5}
 & Clips (\%) & GCC@0.1 & Clips (\%) & GCC@0.1 \\
\midrule
Ped1 & 72.5 & $47.53_{\pm 0.08}$ & 27.5 & $39.39_{\pm 0.86}$ \\
Ped2 & 41.7 & $43.33_{\pm 0.94}$ & 58.3 & $37.14_{\pm 0.00}$ \\
Avenue & 54.8 & $41.27_{\pm 0.50}$ & 45.2 & $40.83_{\pm 0.45}$ \\
ShanghaiTech & 55.8 & $31.01_{\pm 0.23}$ & 44.2 & $23.70_{\pm 0.47}$ \\
UCF-Crime & 26.0 & $44.95_{\pm 0.82}$ & 74.0 & $27.89_{\pm 0.15}$ \\
MSAD & 38.3 & $43.19_{\pm 0.23}$ & 61.7 & $24.89_{\pm 0.23}$ \\
\midrule
Pooled & 43.4 & $40.10_{\pm 0.30}$ & 56.6 & $27.91_{\pm 0.25}$ \\
\bottomrule
\end{tabular}
\end{table*}

\FloatBarrier

\subsection{Participant Recall and Extra Predictions}
\label{sec:supp_coverage}

Table~\ref{tab:supp_coverage} complements the joint scores with participant recall and the number of unmatched predicted tracks.
Recall uses an IoU threshold of 0.1 and positive pixel intersection.
FPI is averaged over video clips.

\begin{table*}[t]
\centering
\caption{Participant recall and unmatched predicted tracks.}
\label{tab:supp_coverage}
\footnotesize
\setlength{\tabcolsep}{3pt}
\renewcommand{\arraystretch}{1.1}
\begin{tabular}{lcccccc}
\toprule
Model & \multicolumn{2}{c}{\textbf{Anomaly}} & \multicolumn{2}{c}{\textbf{Appearance}} & \multicolumn{2}{c}{\textbf{Motion}} \\
\cmidrule(lr){2-3} \cmidrule(lr){4-5} \cmidrule(lr){6-7}
 & Recall & FPI & Recall & FPI & Recall & FPI \\
\midrule
\multicolumn{7}{l}{\textit{Joint caption and mask generation}} \\
Sa2VA-8B & 47.77 & 0.107 & 42.07 & 0.449 & 48.23 & 0.806 \\
VideoGLaMM-3.8B & 32.44 & 0.391 & 28.05 & 1.223 & 30.93 & 1.148 \\
UniPixel-7B & 11.76 & 0.100 & 3.67 & 0.097 & 18.74 & 1.036 \\
\midrule
\multicolumn{7}{l}{\textit{Captioning with separate localization}} \\
Veason-R1-3B & 53.65 & 0.898 & 33.55 & 0.461 & 34.86 & 0.638 \\
InternVL3-8B + SAM2 & 53.65 & 0.575 & 42.20 & 1.032 & 43.12 & 1.199 \\
Qwen3-VL-8B + SAM2 & 72.01 & 0.464 & 54.52 & 0.425 & 55.70 & 0.444 \\
InternVL3.5-8B + SAM2 & 38.50 & 0.502 & 45.87 & 1.400 & 40.37 & 1.573 \\
Qwen3.5-9B + SAM2 & 71.12 & 0.510 & 66.97 & 1.403 & 66.06 & 1.515 \\
\bottomrule
\end{tabular}
\end{table*}

\FloatBarrier

\subsection{Role Classification with Known Participants}
\label{sec:supp_role_classification}

All four VLMs receive the same marked frames for each clip, with up to eight frames sampled from the anomaly window.
Contours follow the ground-truth masks, while instance numbers and colors are shuffled independently for each clip.
The inputs contain no ground-truth role assignments, captions, or textual instance descriptions.
Each model selects the instance numbers for the acting and affected roles using greedy decoding.
This experiment evaluates role assignment with participant locations supplied.

The models were run on all benchmark clips.
The role-classification comparison uses clips in which both roles appear in the sampled frames, covering 35.9\% of the benchmark.
Within this subset, 50.3\% of marked instances are acting and 49.7\% are affected.
Only instances visible in the sampled frames receive a number for role classification.

Acting and affected accuracy pool correct assignments over the instances of each ground-truth role.
Exact match is the percentage of clips in which every marked participant receives the correct role.
An instance assigned to both roles or neither role counts as incorrect.
Invalid responses are scored as empty assignments.

Table~\ref{tab:role_classification} reports accuracy for each role and exact match over all marked participants.
The Qwen models have similar acting and affected accuracy when participant locations are supplied.
However, neither model assigns every participant correctly in a majority of these clips.

\begin{table}[t]
\centering
\caption{Role classification with participant locations supplied.}
\label{tab:role_classification}
\small
\setlength{\tabcolsep}{3pt}
\renewcommand{\arraystretch}{1.1}
\begin{tabular}{lrrr}
\toprule
Model & Acting & Affected & Exact match \\
\midrule
InternVL3-8B & 64.4 & 33.9 & 31.8 \\
InternVL3.5-8B & 49.2 & 54.0 & 35.8 \\
Qwen3-VL-8B & 63.4 & 61.9 & 47.3 \\
Qwen3.5-9B & 64.9 & 60.3 & 49.3 \\
\bottomrule
\end{tabular}
\end{table}

\begin{figure*}[t]
  \centering
  \includegraphics[width=\textwidth]{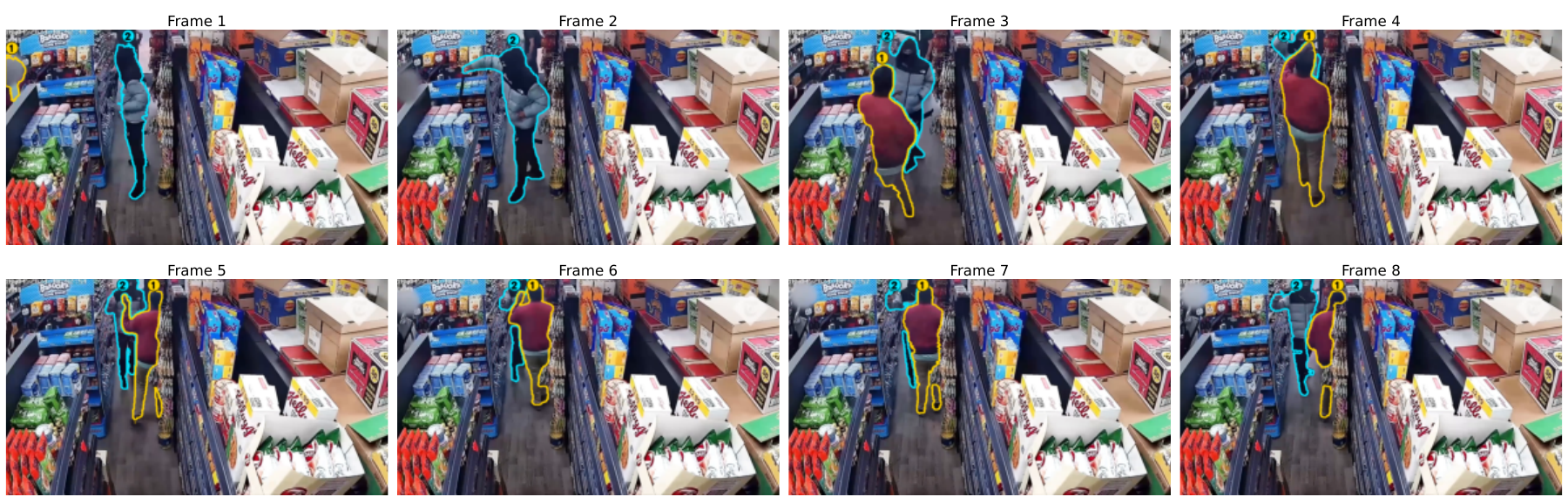}
  \caption{Input frames for role classification. Numbers and colors identify instances without indicating their roles.}
  \label{fig:role_classification_inputs}
\end{figure*}

Figure~\ref{fig:role_classification_inputs} shows an input example from MSAD.
In this clip, instance 2 is acting and instance 1 is affected.
Correctly labeling instance 2 while omitting instance 1 does not count as an exact match.
The prompt below uses the same instance numbers as Figure~\ref{fig:role_classification_inputs}.

\paragraph{Role classification prompt.}
\begin{quote}
\small\ttfamily
These are 8 frames in time order from a surveillance video that contains an anomalous event. Some people or objects are outlined, and each one has a number label. The same number marks the same person or object in every frame. The numbers in this video are: 1, 2.
Which numbered people or objects perform the anomalous action (acting), and which are affected by it (affected)? A number can be in neither list. The affected list can be empty. Answer only with JSON: \{"acting": [numbers], "affected": [numbers]\}

\end{quote}

Table~\ref{tab:supp_role_confusion} reports the confusion matrices as percentages within each ground-truth role.
Other includes instances assigned to neither role or to both roles, including those with invalid responses.
InternVL3 assigns 46.6\% of affected instances to the acting role.
The corresponding rate for Qwen3.5 is 19.0\%, showing that the pattern of role errors varies across models.

\begin{table}[t]
\centering
\caption{Role classification confusion matrices.}
\label{tab:supp_role_confusion}
\footnotesize
\setlength{\tabcolsep}{3pt}
\renewcommand{\arraystretch}{1.1}
\begin{tabular}{llrrr}
\toprule
Model & Ground-truth role & \multicolumn{3}{c}{Predicted role} \\
\cmidrule(lr){3-5}
 & & Acting & Affected & Other \\
\midrule
InternVL3-8B & Acting & 64.4 & 19.9 & 15.7 \\
 & Affected & 46.6 & 33.9 & 19.6 \\
\midrule
InternVL3.5-8B & Acting & 49.2 & 38.2 & 12.6 \\
 & Affected & 31.2 & 54.0 & 14.8 \\
\midrule
Qwen3-VL-8B & Acting & 63.4 & 30.9 & 5.8 \\
 & Affected & 23.3 & 61.9 & 14.8 \\
\midrule
Qwen3.5-9B & Acting & 64.9 & 18.3 & 16.8 \\
 & Affected & 19.0 & 60.3 & 20.6 \\
\bottomrule
\end{tabular}
\end{table}

\FloatBarrier

\subsection{Relative Participant Size}
\label{sec:supp_relative_area}

We examine Qwen3-VL+SAM2 localization by relative image area on the same clip subset used for role classification (Sec.~\ref{sec:supp_role_classification}).
For each role, we sum the image fractions covered by its instance masks and average over the sampled input frames.
Within this subset, the affected group occupies less image area than the acting group in 68.2\% of clips.
In the remaining 31.8\%, its area is at least as large.

Table~\ref{tab:supp_relative_area} reports acting and affected mIoU over every annotated frame of the anomaly clip using the same unordered matching as the main evaluation.
The acting and affected scores are 60.40 and 32.42 when the affected group is smaller, compared with 43.57 and 50.18 in the remaining clips.
The localization gap is therefore concentrated in events with smaller affected regions for this pipeline.
This comparison describes an association with relative image area, as the groups also differ in scene and event composition.

\begin{table}[t]
\centering
\caption{Qwen3-VL+SAM2 localization by relative image area.}
\label{tab:supp_relative_area}
\small
\setlength{\tabcolsep}{3pt}
\renewcommand{\arraystretch}{1.1}
\begin{tabular}{lrrr}
\toprule
Affected area & Clips (\%) & Acting & Affected \\
\midrule
Smaller & 68.2 & 60.40 & 32.42 \\
Not smaller & 31.8 & 43.57 & 50.18 \\
\bottomrule
\end{tabular}
\end{table}

\FloatBarrier

\section{Broader Impacts and Release}
\label{sec:supp_responsible_release}

\FloatBarrier

\subsection{Broader Impacts}
\label{sec:supp_broader_impacts}

AnomGround provides instance masks, role labels, captions, and text-to-track links for evaluating whether anomaly explanations are supported by visual evidence.
Joint evaluation can reveal errors missed by caption similarity or mask overlap alone.

Detailed surveillance annotations can also make VAD systems easier to train and deploy.
Generated explanations can mislead users when applied to decisions about individuals without human review.
Acting and affected labels describe visible participation in dataset-annotated events.
They do not establish intent, blame, danger, or legal responsibility.
The benchmark is intended for research evaluation, and its results do not establish suitability for operational decisions about individuals.

\FloatBarrier

\subsection{Release Plan}
\label{sec:supp_release_plan}

Upon acceptance, we will release the benchmark annotations (captions, role labels, mask tracks, and text-to-track links) and evaluation code.
For datasets whose raw videos cannot be redistributed, we will release sidecar annotations linked by video identifiers and instructions to reproduce all results using videos obtained from the official sources under their licenses.
We will also provide a datasheet describing formats, splits, and licensing notes.

\end{document}